\documentclass[journal]{IEEEtai}

\usepackage[colorlinks,urlcolor=blue,linkcolor=blue,citecolor=blue]{hyperref}

\usepackage{color,array}

\usepackage{graphicx}

\usepackage{amsmath,amsfonts}
\usepackage{algorithmic}
\usepackage{algorithm}
\usepackage{multirow}
\usepackage{array}
\usepackage[caption=false,font=normalsize,labelfont=sf,textfont=sf]{subfig}
\usepackage{textcomp}
\usepackage{stfloats}
\usepackage{url}
\usepackage{booktabs}
\usepackage{verbatim}
\usepackage{cite}

\usepackage[labelsep=period]{caption}
\usepackage[T1]{fontenc}

\graphicspath{{../Figures/}}
\DeclareGraphicsExtensions{.pdf,.png,.jpeg}

\usepackage[flushleft]{threeparttable}

\newcommand{\includepdftrim}[1]{\includegraphics[width=0.45\columnwidth,trim=140 250 140 250]{#1}}

\newcommand{\includehist}[1]{\includegraphics[width=\columnwidth,trim=90 0 50 0,clip]{#1}}
\usepackage{soul}
\newtheorem{thm}{Theorem}[section]
\newtheorem{lem}[thm]{Lemma}

\usepackage{xcolor}

\begin{document}

\title{Soft-Constrained Optimization of Latent Space in Variational Autoencoders}

\author{Ye~Shi,~\IEEEmembership{Student~Member,~IEEE}
\thanks{The authors are with the Elmore Family School of Electrical and Computer Engineering, Purdue University, West Lafayette, IN, 47906 USA (e-mail: shiye@purdue.edu)}
\thanks{Manuscript received TBD; revised TBD}}

\markboth{IEEE Transactions on Artificial Intelligence, Vol. 00, No. 0, Month 2026}
{Shi: Soft-Constrained Optimization of Latent Space in Variational Autoencoders}

\maketitle

\begin{abstract}
The variational autoencoder (VAE) is an unsupervised neural-network structure
that has been adapted to a wide range of learning tasks, and its usefulness
depends on two properties of its latent space that are difficult to obtain
simultaneously: high encoding capacity in the individual latent variables and a
low-dimensional, disentangled organization of those variables.
Weakening the Kullback--Leibler (KL) regularization raises capacity but degrades
disentanglement, whereas strengthening it improves disentanglement but prunes
latent variables away entirely.
This paper formulates VAE training as a soft-constrained optimization problem and
introduces two components that target the two facets of this trade-off.
First, we impose an entropy-based constraint (EC) on individual latent variables.
We show that, for the generative factors underlying the data, the entropy of a
latent code upper-bounds the mutual information it can carry about those factors,
and that the bound is attained in the deterministic-encoder limit; raising the
entropy therefore raises the capacity available for encoding.
Second, we propose a weight-filter method that exploits the slack of the soft
constraint to prune low-entropy latent dimensions during downstream training,
which yields an empirical lower bound on the usable latent-space dimensionality.
On dSprites, the EC increases the aggregate latent-variable activation score by
43--62\% relative to a vanilla VAE and by up to 17\% relative to a $\beta$-VAE,
improves our correlation-based disentanglement score by 9--14\%, and attains the
highest FactorVAE score among the $\beta$-VAE variants ($0.891$ versus $0.847$);
the best configuration also lowers the reconstruction error by 38\%, although
this reduction is not uniform across configurations.
On MNIST, the weight filter reduces the latent dimensionality supplied to a
downstream classifier from ten to two while holding classifier accuracy above
$90\%$, and reaches convergence in 37\% fewer epochs at that dimensionality than
the same procedure without the EC.
Finally, we characterize how the two kinds of generative factors are encoded:
low-entropy discrete factors tend to be merged into a single latent variable,
whereas high-entropy continuous factors are distributed across several.
\end{abstract}

\begin{IEEEImpStatement}
Unsupervised representation learning is used wherever labels are scarce, but the
representations that autoencoders learn are often either too entangled to
interpret or larger than the downstream task requires.
This work addresses both problems within a single soft-constrained formulation.
The entropy-based constraint gives practitioners a principled quantity---the
entropy of each latent variable---to monitor and control, rather than a single
regularization weight that trades capacity against disentanglement.
The weight-filter method identifies, during training and without a separate
search, how many latent dimensions a downstream task actually needs; in our
MNIST classification study it reduced that number from ten to two while holding
accuracy above the specified threshold.
Smaller latent spaces reduce the cost of the downstream models that consume
them and make the learned factors easier to inspect, which matters in settings
where a representation must be audited before it is trusted.
The analysis of how discrete and continuous generative factors are encoded also
gives practitioners a diagnostic for deciding whether a latent space of a given
size can represent the factors present in their data.
\end{IEEEImpStatement}


\section{Introduction}
\IEEEPARstart{V}{ariational} Autoencoder (VAE) is a powerful and flexible 
unsupervised neural-network structure that is suitable for a wide range of applications 
such as image processing, natural language, and robotics~\cite{Kingma_Welling_2019}. 
Although it was initially introduced as a generative model~\cite{Kingma_Welling_2014}, 
growing research in VAE has shown that it is more suitable for 
latent-space-based tasks such as dimensionality compression\cite{theis2017lossy,Rolinek_Zietlow_Martius_2019}, representation learning~\cite{Chen_Li_Grosse_Duvenaud_2019, Higgins_Amos_Pfau_Racaniere_Matthey_Rezende_Lerchner_2018,Eastwood_Williams_2018, bouchacourt2017multi, Locatello2020}, and anomaly detection~\cite{parkMultimodalAnomalyDetector2018,marimontAnomalyDetectionLatent2020,arvanitidisLatentSpaceOddity2021}. 

The latent space of VAEs is a crucial component structure 
that enables it to compress high-dimensional input data into a low-dimensional space.
The latent space is represented by a set of latent variables that abstract the
data features of the training set and that are independent under the factorized
prior, although the aggregate posterior they induce need not be.

Several studies have identified the over-pruning problem of latent variables 
in VAE models~\cite{burdaImportanceWeightedAutoencoders2016, yeungTacklingOverpruningVariational2017, aspertiSparsityVariationalAutoencoders2019}. 
While training a VAE model with sufficient latent dimensions, 
some of the latent variables may fail to encode any data features,
resulting in poor downstream performance. 
This problem is caused by a strong regularization term on the Kullback-Leibler (KL) divergence, 
which encourages the latent space to be as close to a standard normal distribution as possible.
To overcome the over-pruning problem, 
researchers have used different techniques to weaken the regularization 
such as alternating the estimation of the KL divergence term~\cite{burdaImportanceWeightedAutoencoders2016,tolstikhinWassersteinAutoEncoders2019a,aspertiSparsityVariationalAutoencoders2019} and masking the latent space~\cite{yeungTacklingOverpruningVariational2017,khanEpitomicVariationalGraph2020}. 
A weaker KL divergence term may encourage the latent space to encode more data. 
However, this relaxation tends to result in a compromised data representation concerning disentanglement and abstraction~\cite{higgins2017beta,aspertiSparsityVariationalAutoencoders2019}.
Nonetheless, with the emergence of new generative models such as Generative Adversarial Networks (GANs)~\cite{goodfellow2014generative} and Diffusion Models~\cite{caoSurveyGenerativeDiffusion2022}, the over-pruning problem has received less attention in recent years. 
It remains a crucial challenge for VAEs to determine 
what constitutes the optimal latent space.

An optimal latent space is a representation of data that maximizes the ability of a model 
to capture and utilize important features of the data. 
It should be able to disentangle the different factors of variation 
that contribute to the data such as shape, color, texture, and style, 
and represent them in a way that allows easy manipulation in downstream tasks. 
An optimal latent space should also be efficient; that is, it should be the
lowest-dimensional representation that still captures the information the
downstream task requires.

We address the optimal latent space problem 
using constrained optimization techniques in two directions -- 
optimizing the capacity of latent variables and the dimensionality of latent space.
The capacity of latent variables is defined as 
the ability to completely encode a data feature and disentangle it from other data features,
allowing the latent variables to preserve more information and less correlation.
We propose an entropy-based constraint (EC) on latent variables to maximize the
capacity of latent variables.
A higher entropy indicates that the latent variable takes a more diverse range of
values and can therefore encode more information.
We show in Section~\ref{sec:formulation} that the entropy of the latent code
upper-bounds the mutual information it carries about the generative factors of
the data, and that the bound is attained when the encoder becomes deterministic.
Maximizing entropy therefore maximizes the capacity available for encoding, and
imposing the EC makes the latent space more expressive.

The dimensionality of latent space refers to the number of latent variables 
used to represent the input data. 
A higher-dimensional latent space can capture more intricate representations of the data features, 
but it may result in increasing the risk of overfitting, 
leading to computational inefficiencies~\cite{aspertiSurveyVariationalAutoencoders2021}. 
To tackle the challenge of determining the optimal dimensionality, 
we introduce a weight-filter method grounded in the soft-constraint property. 
The method searches for the smallest latent-space dimensionality that a given
downstream task tolerates, and thereby yields an empirical lower bound on that
dimensionality.
Unlike hard constraints, soft constraints can be loosely violated, 
but there is a penalty associated with each violation. 
Our proposed approach eliminates latent variables that do not satisfy the EC well, 
resulting in a shrunken latent space that better balances the trade-off 
between {\em expressiveness} and {\em efficiency}.

In our study, we have observed two properties of latent variables in VAEs:
\begin{itemize}
\item Continuous data features with high entropy, such as the $X$ and $Y$
positions, tend to be encoded in a more disentangled manner, and a single feature
may be distributed across more than one latent variable.
\item When data features are discrete and take few values, such as shape and
scale, a single latent variable may encode several of them jointly. The
combined feature then has a higher entropy than either feature alone, which is
what the constraint rewards.
\end{itemize}

Our primary contributions are as follows.
\begin{enumerate}
\item We cast VAE training as a soft-constrained optimization problem in which
reconstruction accuracy, KL regularization, and latent-variable entropy appear as
separate constraints with separate multipliers, which makes the capacity
requirement explicit rather than implicit in a single regularization weight.
\item We introduce an entropy-based constraint on individual latent variables and
relate it to the mutual information between the latent code and the generative
factors of the data: the entropy upper-bounds that mutual information, with
equality in the deterministic-encoder limit. Empirically, the constraint raises
the aggregate activation score of the latent space by 43--62\% over a vanilla VAE
and lowers the reconstruction error by up to 38\%, while keeping disentanglement
scores competitive.
\item We propose a weight-filter method that prunes low-entropy latent dimensions
during downstream training. On MNIST it reduces the latent dimensionality
consumed by the classifier from ten to two---an 80\% reduction---subject to an
accuracy threshold, and identifies two as the point at which the threshold can no
longer be met.
\item We characterize the encoding behavior of latent variables as a function of
the entropy of the underlying data feature: high-entropy continuous features
(spatial coordinates) are encoded in a disentangled manner and may be split
across latent variables, whereas low-entropy discrete features (shape, scale) are
merged into a single latent variable.
\end{enumerate}

The paper is organized as follows.
Section~\ref{sec:relatedwork} presents an overview of related work 
and provides introductory concept about constrained optimization in VAEs.
Section~\ref{sec:approach} outlines the optimal latent space problem and introduces our proposed approach, which is based on soft-constrained optimization.
Section~\ref{sec:experiment} presents the experimental results derived from our approach.
Section~\ref{sec:discussion} provides a detailed analysis of the distributions of the latent variables. 
Finally, Section~\ref{sec:conclusion} concludes the paper by summarizing the impacts of our proposed approach and suggesting potential directions for future research.

\section{Related Work and Preliminaries}\label{sec:relatedwork}
Latent-space optimization can take various forms depending on the task and the definition of optimality. 
Our approach is situated within the broader literature 
on VAE capacity optimization and optimization of latent-space dimensionality, 
and it draws inspiration from constrained optimization in deep learning. 
We first briefly review these areas and their connection to our proposed work, 
and we then provide an overview of the preliminaries for a better understanding 
of the constrained optimization in VAEs.

\subsection{VAE Capacity Optimization}
In VAEs, the problem of over-pruning poses a trade-off 
between downstream performance and disentangled representation, 
which leads to two directions in optimizing VAE capacity.

\begin{enumerate}
\item
{\em Increasing model capacity by weakening the regularization term.}
To enhance the downstream performance, 
this approach involves weakening the regularization term to increase the encoding capacity of latent space. 
For example, IWAE~\cite{burdaImportanceWeightedAutoencoders2016} estimates the KL divergence 
through weighted importance using the Monte-Carlo method, 
WAE~\cite{tolstikhinWassersteinAutoEncoders2019a} replaces the KL divergence 
with the Wasserstein distance, and eVAEs~\cite{khanEpitomicVariationalGraph2020,yeungTacklingOverpruningVariational2017} regulates 
the KL divergence of contiguous subsets of the latent space by introducing uniformly-distributed masks. 
Another approach is to find alternative closed-form expressions for the KL divergence 
such as ~\cite{aspertiSparsityVariationalAutoencoders2019} and~\cite{aspertiBalancingReconstructionError2020}.
Many researchers~\cite{burdaImportanceWeightedAutoencoders2016,yeungTacklingOverpruningVariational2017,cemgilAutoencodingVariationalAutoencoder2020,sinhaConsistencyRegularizationVariational2022} observed more latent variables become active when enhancing the downstream tasks.
The latent-variable activation score (LVAS)~\cite{burdaImportanceWeightedAutoencoders2016} is a common metric for judging whether a latent variable actively encodes data, and hence for counting how many latent variables have been over-pruned.

\item
{\em Regularizing the model to encourage disentangled representation.}
Disentangled representation aims to establish a one-to-one correspondence 
between data features and codec variables through unsupervised learning, 
mimicking human reasoning in summarizing high-dimensional data in low-dimensional features. 
The disentanglement approach seeks to encode each data feature into a separate latent variable independently.
For example, one common technique is to use the $\beta$-VAE approach~\cite{higgins2017beta}, 
which introduces a hyper-parameter $\beta$ to control the balance between the reconstruction error and the disentanglement constraint in the loss function. 
Another approach is to use adversarial training methods~\cite{Chen_Li_Grosse_Duvenaud_2019} 
or other regularization methods such as InfoGAN~\cite{InfoGAN_2016} or FactorVAE~\cite{Kim_Mnih} 
to encourage the model to learn a more disentangled representation.
However, researchers have yet to agree on a common metric. 
There are several metrics for measuring and quantifying disentanglement, 
and they are mostly based on approximating the correlation and covariance 
between feature variables 
and latent variables~\cite{Eastwood_Williams_2018,Locatello2020,higgins2017beta,Kim_Mnih}. 
Recently, researchers~\cite{horanWhenUnsupervisedDisentanglement2021,geEncouragingDisentangledConvex2023,khemakhemVariationalAutoencodersNonlinear2020} prefer to use the correlation to evaluate the disentanglement.
\end{enumerate}

Both directions have their advantages and drawbacks, 
and the choice of the approach may depend on the specific task and the dataset at hand. 
The challenge is to find a balance between generative performance and disentanglement in the latent space, which can lead to more effective and interpretable models.

\subsection{Dimensionality of Latent Space}
The dimensionality of latent space is a crucial hyper-parameter in VAE-based models 
and is often associated with dimensionality reduction.
The VAE structure has demonstrated an exceptional ability for dimensionality reduction 
in various types of data~\cite{lopezGDVAEsGeometricDynamic2022,baggenstossNonlinearDimensionReduction2022,kaurVariationalAutoencoderBasedDimensionality2021}.
Some analyses~\cite{JMLR:v19:17-704, Rolinek_Zietlow_Martius_2019} have revealed 
that VAEs work as non-linear principal component analysis~(PCA).
Unlike PCA, however, the capacity of each latent variable is limited by the
Gaussian prior it is regularized towards: reducing the number of latent
dimensions does not, by itself, give the surviving dimensions more capacity.
Hence, while dimensionality reduction simplifies the representation and improves
computational efficiency for a given task, it may also degrade reconstruction
quality and downstream-task performance.

Conversely, if the latent-space dimension of a VAE is too high, 
it can lead to poor generalization performance on unseen data 
because the VAE may not be able to capture the true distribution of the input data.
However, compared to GANs, the optimal dimensionality of latent space in VAEs 
has received less attention and discussion.
Previous studies on GANs have mainly focused on optimizing 
the dimensionality of latent space using elbow methods linked to the output~\cite{Tripp_2020, bojanowski2017optimizing, Mondal_Chowdhury_Jayendran_Asnani_Singla_P_2020}. 
However, these methods are not applicable to VAEs because the latent space cannot be directly manipulated in the same way.

\subsection{Constrained Optimization in Deep Learning}
Constraints are commonly used in deep learning as a form of regularization for 
stochastic gradient descent (SGD) algorithms~\cite{goodfellow2016deep}. 
By limiting the feasible search space, the constraints help to prevent overfitting, improve generalization, 
and accelerate convergence~\cite{ravi2019explicitly,pathak2015constrained,achiam2017constrained}. 
These implicit constraints are typically designed according to guidelines 
such as $L_p$-norms~\cite{Yoshida_Miyato_2017,ayinde2017deep,Wu_Li_Deng_Liu_Wu_Xie_Shi_2019}, 
prior probability~\cite{higgins2017beta}, Markov decision process~\cite{achiam2017constrained}, 
and indicator functions~\cite{zhang2018tau,ravi2018constrained,ravi2019explicitly}.

Constrained optimization is often applied to the training of autoencoders~\cite{Kingma_Welling_2014,Kim_Mnih,higgins2017beta}, 
which have been shown to enhance their ability to comprehend and disentangle latent representations~\cite{ghiassirad2019application,ayinde2017deep}.
Recent research has explored the use of deep learning to solve constrained optimization problems~\cite{Kotary_Fioretto_Van_Hentenryck_Wilder_2021}. 
Autoencoders have shown significant potential in constrained optimization 
by transforming the original data from a discrete space to a continuous latent space~\cite{bentleyCOILConstrainedOptimization2022}.

\subsection{Constrained Optimization in VAEs}\label{sec:VAE}
The vanilla autoencoder optimizes the parameters $\phi$ and $\theta$ 
of the encoder and decoder, respectively, assuming that
\begin{equation}
\label{eqn:vae_assumption}
q_\phi(\mathbf{z}|\mathbf{x})\approx p_\theta(\mathbf{z}|\hat{\mathbf{x}}),
\end{equation}
where $q_\phi(\mathbf{z}| \mathbf{x})$ is the approximate posterior probability
density function~(PDF) of the latent vector $\mathbf{z}$ produced by the encoder,
$p_\theta(\mathbf{z} | \hat{\mathbf{x}})$ is the posterior PDF of $\mathbf{z}$
implied by the decoder, $\mathbf{x}$ is the input, and $\hat{\mathbf{x}}$ is the
output, with $\hat{\mathbf{x}}\approx \mathbf{x}$.

VAE maximizes the marginal likelihood of $p_\theta(\mathbf{x})$ 
and expands it by Eq.~\eqref{eqn:vae_assumption} as
\begin{equation}
\label{eqn:p_theta_x}
\begin{aligned}
       \ln{p_\theta(\mathbf{x})} =& \mathbb{E}_{q_{\phi}(\mathbf{z} |\mathbf{x})}\left[\ln \left[\frac{p_{\boldsymbol{\theta}}(\hat{\mathbf{x}}, \mathbf{z})}{q_{\phi}(\mathbf{z}  |\mathbf{x})}\right]\right]\\
&+
\mathrm{D}_{\mbox{KL}}\left(q_{\phi}(\mathbf{z} | \mathbf{x}) \| p_{\theta}(\mathbf{z}| \hat{\mathbf{x}})\right),
\end{aligned}
\end{equation}
where $\mathbb{E}\left(\cdot\right)$ is the expectation operator
and $\mathrm{D}_{\mbox{KL}}\left(\cdot\|\cdot\right)$ is the KL divergence function.
Under the assumption in Eq.~\eqref{eqn:vae_assumption}, 
the second term on the right-hand side of Eq.~\eqref{eqn:p_theta_x} is the KL divergence of $q_{\phi}(\mathbf{z}| \mathbf{x})$ from $p_{\theta}(\mathbf{z}| \hat{\mathbf{x}})$, which is non-negative,
\begin{equation}
    \mathrm{D}_{\mbox{KL}}\left(q_{\phi}(\mathbf{z} | \mathbf{x}) \| p_{\theta}(\mathbf{z}| \hat{\mathbf{x}})\right)\geq 0.
\end{equation}
The first term on the right-hand side of Eq.~\eqref{eqn:p_theta_x} then is an evidence lower bound (ELBO) of $\ln{p_{\theta}(\mathbf{x})}$ as
\begin{equation}
\label{eqn:VAElossfunc}
\begin{aligned}
  \mathcal{L}_{V}(\theta, \phi ; \mathbf{x}, \hat{\mathbf{x}}, \mathbf{z})
  =&\mathbb{E}_{q_{\phi}(\mathbf{z} |\mathbf{x})}\left[\ln \left[\frac{p_{\boldsymbol{\theta}}(\hat{\mathbf{x}}, \mathbf{z})}{q_{\phi}(\mathbf{z}  |\mathbf{x})}\right]\right]\\
  =&\mathbb{E}_{q_{\phi}(\mathbf{z} | \mathbf{x})}\left[\ln p_{\theta}(\hat{\mathbf{x}} | \mathbf{z})\right] \\ 
  &-\mathrm{D}_{\mbox{KL}}\left(q_{\phi}(\mathbf{z} | \mathbf{x}) \| p(\mathbf{z})\right),
\end{aligned}
\end{equation}
where $p(\mathbf{z})$ is the PDF of $\mathbf{z}$. 
Equation~\eqref{eqn:VAElossfunc} then becomes the objective function of VAE 
because it is with respect to the parameters $\theta$ and $\phi$, 
and maximizing it leads to a higher marginal likelihood of $p_\theta(\mathbf{x})$ 
and minimizes the KL divergence of $q_{\phi}(\mathbf{z}| \mathbf{x})$ from $p(\mathbf{z})$.

Because $p(\mathbf{z})$ is difficult to estimate during training,
VAE introduces a new random variable, $\xi$, with a known PDF, $p(\xi)$,
to formulate Eq.~\eqref{eqn:VAElossfunc} as a hard-constrained optimization,
\begin{equation}
    \begin{aligned}
\label{eqn:VAEopt}
    \max_{\phi, \theta}& \quad\mathcal{L}_{V}(\theta, \phi ; \mathbf{x}, \hat{\mathbf{x}},\mathbf{z})\\
     \textrm{s.t.} &\quad \mathbf{z} =  g_\theta(\xi,\mathbf{x}), \quad \mbox{where }\xi\sim p\left(\xi\right),
\end{aligned}
\end{equation}
where $g_{\theta}(\cdot)$ is a mapping function for the reparameterization trick~\cite{Devroye_1986} .

\begin{figure}[tb]
\centering
\includegraphics[width=0.45\textwidth]{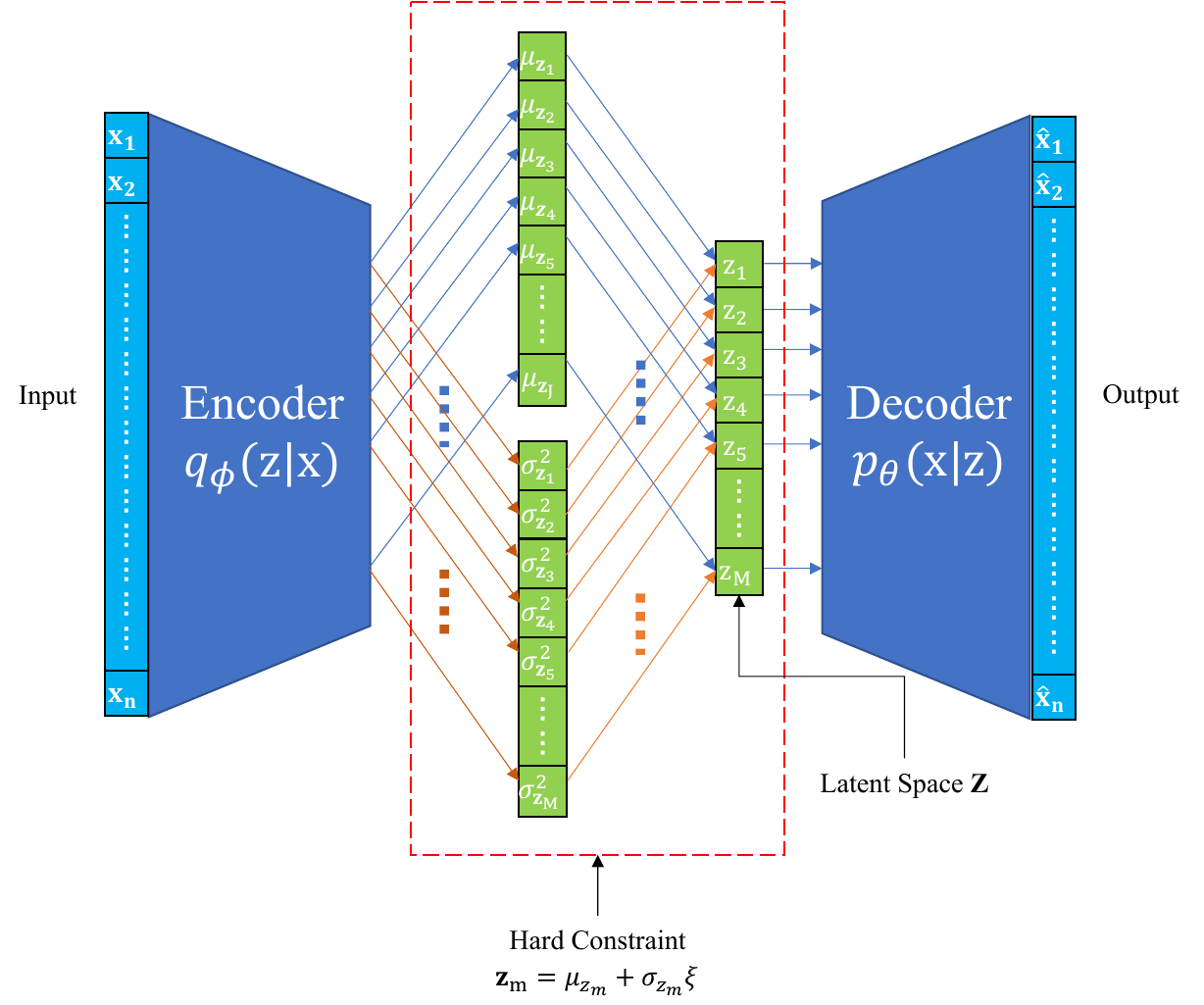}
\caption{VAE structure.
The red dashed box demonstrates 
how the latent variable maps to a random variable in a normal distribution.}
\label{fig:VAE}
\end{figure}

The vanilla VAE embeds the hard constraint in its structure as shown in Fig.~\ref{fig:VAE}.
Each latent variable $\mathbf{z}_m$ is mapped 
by $\mathbf{z}_m = \mu_{\mathbf{z}_m}+\sigma_{\mathbf{z}_m}\mathcal{\xi}$
and $\mathcal{\xi}\sim \mathcal{N}(0,1)$, 
where $\mathbf{z}_m$ is a latent variable in the latent vector $\mathbf{z}$,
$M$ is the dimension of the latent space,
$m$ is an arbitrary integer, $m\in\left[1,M\right]$,
$\mu_{\mathbf{z}_m}$ is the mean of $\mathbf{z}_m$, 
and $\sigma_{\mathbf{z}_m}$ is the standard deviation of $\mathbf{z}_m$.
Because this construction makes $q_\phi(\mathbf{z}|\mathbf{x})$ a diagonal
Gaussian, a closed-form expression of the KL divergence with respect to the
standard normal prior is obtained as
\begin{equation}
\label{eqn:DKLanalytic}
\begin{aligned}
    \mathrm{D}_{\mbox{KL}}\left(q_{\phi}(\mathbf{z} | \mathbf{x}) \| p(\mathbf{z})\right) 
    = & -\frac{1}{2}\sum_{m=1}^{M}\left(1+\ln{\sigma^2_{\mathbf{z}_m}}-
    \mu_{\mathbf{z}_m}^2-\sigma^2_{\mathbf{z}_m}\right).\\
\end{aligned}
\end{equation}

$\beta$-VAE~\cite{higgins2017beta} uses the vanilla VAE structure 
and decomposes the ELBO in Eq.~\eqref{eqn:VAElossfunc} by separating 
the first term, $\mathbb{E}_{q_{\phi}(\mathbf{z} | \mathbf{x})}\left[\ln p_{\theta}(\hat{\mathbf{x}} | \mathbf{z}))\right]$, as the reconstruction loss, 
and the second term, $\mathrm{D}_{\mbox{KL}}\left(q_{\phi}(\mathbf{z} | \mathbf{x}) \| p(\mathbf{z})\right)$, 
as the regularization that encourages disentangled latent variables.
Hence, $\beta$-VAE formulates the constrained optimization as
\begin{equation}
\begin{aligned}
\label{eqn:beta-vae1}
    \max_{\phi, \theta}&{\quad \mathbb{E}_{q_{\phi}(\mathbf{z} | \mathbf{x})}\left[\ln p_{\theta}(\hat{\mathbf{x}} | \mathbf{z})\right]} \quad\\
     \textrm{s.t.} &\quad \mathrm{D}_{\mbox{KL}}\left(q_{\phi}(\mathbf{z} | \mathbf{x}) \| p(\mathbf{z})\right)<\epsilon,
\end{aligned}
\end{equation}
where $\epsilon$ is a relaxation factor
and $\epsilon\geq0$ holds the complementary slackness condition of the soft constraint.
For the Bernoulli decoder used throughout this paper, maximizing the objective
function is equivalent to minimizing the cross entropy between the input and the
reconstruction.
The constraint pulls the approximate posterior towards the factorized prior and
thereby encourages the latent variables to be mutually independent.

We formulate Eq.~\eqref{eqn:beta-vae1} as a Lagrangian function
under the Karush-Kuhn-Tucker (KKT) conditions~\cite{Karush_2014,Kuhn_Tucker_1951}, 
\begin{equation}
\begin{aligned}
\label{eqn:beta-vae2}
\mathcal{F}_{\beta}(\theta, \phi, \beta ; \mathbf{x}, \hat{\mathbf{x}}, \mathbf{z}) &= \mathbb{E}_{q_{\phi}(\mathbf{z} | \mathbf{x})}\left[\ln p_{\theta}(\hat{\mathbf{x}} | \mathbf{z}))\right]\\ & \quad -\beta\left(\mathrm{D}_{\mbox{KL}}\left(q_{\phi}(\mathbf{z} | \mathbf{x}) \| p(\mathbf{z})\right)-\epsilon\right) ,
\end{aligned}
\end{equation}
where $\beta$ is a saddle point under the KKT conditions.
Due to the dual feasibility condition, $\beta\geq0$, 
$\beta$---referred to as a KKT multiplier---is typically manually assigned due to the complexity of deep learning models. 
Consequently, these constraints operate as penalties within a soft-constrained framework.
In practice, the constraint term is conducted as a summation of 
the KL divergence of all the latent variables.

The $\beta$-VAE loss function is obtained as 
\begin{equation}
  \begin{aligned}
\label{eqn:beta-vae_loss}
  \mathcal{L}_{\beta}(\theta, \phi, \beta ; \mathbf{x}, \hat{\mathbf{x}},  \mathbf{z})
  =&\mathbb{E}_{q_{\phi}(\mathbf{z} | \mathbf{x})}\left[\ln p_{\theta}(\hat{\mathbf{x}} | \mathbf{z}))\right] \\ 
  &-\beta \mathrm{D}_{\mbox{KL}}\left(q_{\phi}(\mathbf{z} | \mathbf{x}) \| p(\mathbf{z})\right).
\end{aligned}
\end{equation}
When $\beta=1$, $\beta$-VAE corresponds to the vanilla VAE. 
Adjusting the value of $\beta$ in $\beta$-VAE can have significant effects 
on the training process and the model performance. 
Increasing the $\beta$ value places more emphasis on the soft constraint, 
resulting in better disentanglement of latent variables 
but poorer reconstruction quality. 
Conversely, decreasing the $\beta$ value improves the generative performance 
but may result in less disentangled latent variables. 
While this approach reduces the divergence between the prior, $p(\mathbf{z})$,
and the approximate posterior, $q_\phi(\mathbf{z}|\mathbf{x})$, it does so at the
cost of the information the latent variables retain about the input, and hence
reduces their capacity~\cite{burgess2018understanding}.
Hence, the reconstruction performance of the $\beta$-VAE decoder 
suffers when the $\beta$ value is increased.
The $\beta$ value is often considered as a balance parameter 
for the over-pruning problem~\cite{aspertiBalancingReconstructionError2020}.

\section{Proposed Approach}\label{sec:approach}
\subsection{Problem Formulation}\label{sec:formulation}
The primary objective of this study is to achieve optimal latent space of a VAE model 
with the perspective of the capacity of latent variables and the dimensionality of latent space 
in classification tasks.

The encoding capacity is characterized by the mutual information between the
generative factors $\mathbf{y}$ that produce the data $\mathbf{x}$ and the latent
variables $\mathbf{z}$ that encode $\mathbf{x}$,
\begin{equation}
\label{eqn:MIdef}
    \mathrm{I}(\mathbf{y}, \mathbf{z}) = \mathrm{H}(\mathbf{y}, \mathbf{z}) - \mathrm{H}(\mathbf{y}|\mathbf{z}) - \mathrm{H}(\mathbf{z}|\mathbf{y}),
\end{equation}
where $\mathrm{I}(\cdot,\cdot)$ denotes mutual information, $\mathrm{H}(\cdot,\cdot)$ denotes the joint entropy, and $\mathrm{H}(\cdot|\cdot)$ denotes conditional entropy.
Throughout, $\mathrm{H}(\mathbf{z})$ denotes the differential entropy of the
aggregate posterior $q_\phi(\mathbf{z})=\mathbb{E}_{\mathbf{x}}\left[q_\phi(\mathbf{z}|\mathbf{x})\right]$.

\begin{lem}\label{lem:CEEZ}
Zero conditional entropy~\cite{coverEntropyRelativeEntropy2005}. $\mathrm{H}(\mathbf{Y}|\mathbf{X})=0$, if and only if the value of $\mathbf{Y}$ is completely determined by the value of $\mathbf{X}$.
\end{lem}

If the encoder were deterministic, $\mathbf{z}=\text{Encoder}(\mathrm{G}(\mathbf{y}))$
would be completely determined by $\mathbf{y}$, Lemma~\ref{lem:CEEZ} would give
$\mathrm{H}(\mathbf{z}|\mathbf{y})=0$, and Eq.~\eqref{eqn:MIdef} would collapse to
$\mathrm{I}(\mathbf{y},\mathbf{z})=\mathrm{H}(\mathbf{z})$.
The encoder of a VAE, however, is stochastic: the reparameterization in
Eq.~\eqref{eqn:VAEopt} draws $\xi$ independently of $\mathbf{x}$, so
$\mathrm{H}(\mathbf{z}|\mathbf{y})\neq 0$ in general and the residual term must be
retained.
Because $\mathbf{x}=\mathrm{G}(\mathbf{y})$ is deterministic, conditioning on
$\mathbf{y}$ and conditioning on $\mathbf{x}$ leave the same conditional law for
$\mathbf{z}$, and for the diagonal Gaussian encoder that conditional entropy is
available in closed form,
\begin{equation}
\label{eqn:Hzy}
    \mathrm{H}(\mathbf{z}|\mathbf{y}) = \mathrm{H}(\mathbf{z}|\mathbf{x})
    = \mathbb{E}_{\mathbf{x}}\left[\sum_{m=1}^{M}\frac{1}{2}\ln\left(2\pi e\,\sigma^2_{\mathbf{z}_m}\right)\right].
\end{equation}
Writing Eq.~\eqref{eqn:MIdef} in the equivalent form
$\mathrm{I}(\mathbf{y},\mathbf{z})=\mathrm{H}(\mathbf{z})-\mathrm{H}(\mathbf{z}|\mathbf{y})$
and substituting Eq.~\eqref{eqn:Hzy} gives
\begin{equation}
\label{eqn:MIdecomp}
        \mathrm{I}(\mathbf{y}, \mathbf{z}) = \mathrm{H}(\mathbf{z}) -
        \mathbb{E}_{\mathbf{x}}\left[\sum_{m=1}^{M}\frac{1}{2}\ln\left(2\pi e\,\sigma^2_{\mathbf{z}_m}\right)\right].
\end{equation}
The second term of Eq.~\eqref{eqn:MIdecomp} is the entropy injected by the
reparameterization noise, and it is exactly the term the KL regularizer controls:
averaging $\mathrm{D}_{\mbox{KL}}(q_\phi(\mathbf{z}|\mathbf{x})\|p(\mathbf{z}))$
over $\mathbf{x}$ yields
$-\mathrm{H}(\mathbf{z}|\mathbf{x})+\frac{M}{2}\ln 2\pi+\frac{1}{2}\sum_{m}\mathbb{E}[\mu^2_{\mathbf{z}_m}+\sigma^2_{\mathbf{z}_m}]$,
which is minimized at $\sigma^2_{\mathbf{z}_m}=1$.
For models trained with a common $\beta$, the noise term is therefore held near
the entropy of the prior and varies little across models, so differences in
encoding capacity are governed by the first term, $\mathrm{H}(\mathbf{z})$.
Maximizing $\mathrm{H}(\mathbf{z})$ subject to the KL constraint consequently
maximizes the capacity available for encoding the generative factors, and in the
deterministic-encoder limit it maximizes $\mathrm{I}(\mathbf{y},\mathbf{z})$
exactly.

The joint entropy is not directly accessible during training, but it is bounded by
the marginal entropies of the individual latent variables,
\begin{equation}
\label{eqn:TC}
    \mathrm{H}(\mathbf{z}) = \sum_{m=1}^{M}\mathrm{H}(\mathbf{z}_m) - \mathrm{TC}(\mathbf{z}),
    \qquad \mathrm{TC}(\mathbf{z})\geq 0,
\end{equation}
where $\mathrm{TC}(\cdot)$ is the total correlation, which vanishes when the
latent variables are mutually independent.
The KL constraint of Eq.~\eqref{eqn:beta-vae1} already drives
$\mathrm{TC}(\mathbf{z})$ towards zero, so raising the individual marginal
entropies raises $\mathrm{H}(\mathbf{z})$ as well.
This is what makes a per-variable constraint a usable surrogate for the joint
objective, and it is why the two goals of high capacity and low correlation need
not conflict.
The capacity of a latent variable $\mathbf{z}_m$ is accordingly measured by its
differential entropy,
\begin{equation}
\label{eqn:infoEtropy}
\mathrm{H}(\mathbf{z}_m)=-\int q_\phi\left(\mathbf{z}_m\right) \ln q_\phi\left(\mathbf{z}_m\right)\, \mathrm{d}\mathbf{z}_m,
\end{equation}
which Algorithm~\ref{alg:EstEntropy} estimates by the plug-in entropy of a
$K$-bin histogram of $\mathbf{z}_m$.
The two quantities are not identical: for bins of width $\Delta$ the discrete
estimate exceeds the differential entropy by approximately $\log(1/\Delta)$, a
constant offset that cancels in comparisons between models but that must be kept
in mind when the estimate is compared against an absolute scale such as the
entropy of the prior. 

The optimal latent space problem then is considered as a VAE model with a sufficiently high-dimensional latent space,
where some latent variables encounter the over-pruning problem 
resulting in a loss of crucial real data features.
To address this issue, we introduce a subspace of the original latent space for
downstream tasks, with $J\leq M$ as the dimensionality of the subspace and
$\mathbf{z}^\prime$ as the corresponding retained latent variables.
Our objective is to use the smallest such subspace that the downstream task
tolerates, while the encoder and decoder are trained to make the retained latent
variables as informative as possible.
This is naturally written as a bi-level problem,
\begin{equation}
\label{eqn:bilevel}
\begin{aligned}
    \min_{J\in[1,M]} \quad & J \\
    \textrm{s.t.} \quad & (\phi^\star,\theta^\star) \in \arg\max_{\phi, \theta} \ \mathrm{H}(\mathbf{z}^\prime) \\
     & a(\phi^\star,\theta^\star,J) \geq \alpha ,
\end{aligned}
\end{equation}
where $\phi$ and $\theta$ denote the parameters of the encoder and the decoder,
respectively, $a(\cdot)$ is the accuracy of the downstream classifier, and
$\alpha$ is a user-specified accuracy threshold.
The inner problem raises the capacity of the latent variables and is addressed by
the entropy-based constraint of Section~\ref{sec:maxentropy}; the outer problem
shrinks the subspace and is addressed by the weight-filter method of
Section~\ref{sec:Optlatent}.
We solve the two greedily and jointly rather than exactly: the dimensionality is
decremented by one whenever the accuracy constraint is still satisfied, which
returns a feasible---not certifiably optimal---value of $J$.

\subsection{Entropy-Based Constraint (EC) of Latent Variables}\label{sec:maxentropy}
In order to maximize the entropy of latent variables, 
we introduce a lower-bound constraint $\delta$ on the entropy of the latent variable.
By imposing this constraint, 
we effectively increase the entropy of the latent variable while preserving the decoder's functionality.
To achieve this, we incorporate the constraint as a soft constraint by applying a penalty term to the loss function, enabling us to jointly optimize both the loss of VAE and the entropy of latent variables.
This constraint is represented as:
\begin{equation}
\begin{gathered}
\label{eqn:EntropyConstraint0}
\sum_{m=1}^{M}\left(\delta -\mathrm{H}\left(\mathbf{z}_m\right) \right)<\epsilon_\mathrm{H},
\end{gathered}
\end{equation}
where $\delta\geq0$ is the target entropy of each latent variable and
$\epsilon_\mathrm{H}\geq0$ is the permitted slack.
A natural reference scale for $\delta$ is $\frac{1}{2}\ln(2\pi e)\approx1.42$~nats,
the differential entropy of the standard normal prior: choosing $\delta$ at or
below this value asks every latent variable to be at least as informative as a
draw from the prior.
We emphasize that $\frac{1}{2}\ln(2\pi e)$ is not an upper bound on
$\mathrm{H}(\mathbf{z}_m)$.
Among all distributions of variance $\sigma^2$ the Gaussian attains the maximum
differential entropy $\frac{1}{2}\ln(2\pi e\sigma^2)$, so an active latent
variable whose aggregate posterior has variance greater than one exceeds that
value; the experiments in Section~\ref{sec:experiment} confirm this.
Equation~\eqref{eqn:EntropyConstraint0} is therefore a lower-bound constraint
only, and it is the aggregate over the $M$ latent variables that is constrained,
so an individual variable may fall below $\delta$ provided the others compensate.

Using Eq.~\eqref{eqn:beta-vae1}, 
the constrained optimization problem is rewritten with 
Eq.~\eqref{eqn:EntropyConstraint0} as an additional constraint,
\begin{equation}
\begin{aligned}
 \label{eqn:newproblemorigin}
    \max_{\phi, \theta}&{\quad \mathbb{E}_{q_{\phi}(\mathbf{z} | \mathbf{x})}\left[\ln p_{\theta}(\hat{\mathbf{x}} | \mathbf{z}))\right]} \quad \\[0.1in]
     \textrm {s.t.}&\quad \mathrm{D}_{\mbox{KL}}\left(q_{\phi}(\mathbf{z} | \mathbf{x}) \| p(\mathbf{z})\right)<\epsilon_\mathrm{K} \\
      &\quad \sum_{m=1}^{M}\left(\delta-   \mathrm{H}(\mathbf{z}_m)\right) < \epsilon_\mathrm{H} .
\end{aligned}
\end{equation}
The two constraints carry separate slack variables because they are not
interchangeable: $\epsilon_\mathrm{K}$ limits how far the approximate posterior
may drift from the prior, whereas $\epsilon_\mathrm{H}$ limits how much total
entropy the latent space may forfeit.
Equation~\eqref{eqn:newproblemorigin} can be formulated as  
a Lagrangian function under the KKT conditions as
\begin{equation}
\begin{aligned}
\label{eqn:newproblemoriginKKT}
      \mathcal{F}_{\mathrm{H}}&(\theta, \phi, \beta, \gamma ; \mathbf{x}, \hat{\mathbf{x}}, \mathbf{z}) = \mathbb{E}_{q_{\phi}(\mathbf{z} |\mathbf{x})}\left[\ln p_{\theta}(\hat{\mathbf{x}} | \mathbf{z}))\right] \\
      & -\beta \left(\mathrm{D}_{\mbox{KL}}\left(q_{\phi}(\mathbf{z}| \mathbf{x}) \| p(\mathbf{z})\right)-\epsilon_\mathrm{K} \right) \\
    & -\gamma\left( \sum_{m=1}^{M}\left(\delta  -\mathrm{H}(\mathbf{z}_m)\right)-\epsilon_\mathrm{H}\right) ,
\end{aligned}
\end{equation}
where $\gamma\geq 0 $ is the KKT multiplier of the EC and holds the corresponding
complementary slackness condition. 

Because $\beta,\gamma,\epsilon_\mathrm{K},\epsilon_\mathrm{H}\geq0$, dropping the
two constant terms $\beta\epsilon_\mathrm{K}$ and $\gamma\epsilon_\mathrm{H}$
yields a lower bound on Eq.~\eqref{eqn:newproblemoriginKKT} whose maximizer is
unchanged, since the dropped terms do not depend on $\phi$ or $\theta$,
\begin{equation}
\begin{aligned}
\label{eqn:newproblemoriginloss}
    \mathcal{L}_{\mathrm{H}}& (\theta, \phi, \beta, \gamma ; \mathbf{x}, \hat{\mathbf{x}}, \mathbf{z}) = \mathbb{E}_{q_{\phi}(\mathbf{z} | \mathbf{x})}\left[\ln p_{\theta}(\hat{\mathbf{x}} | \mathbf{z}))\right] \\
  &  -\beta \mathrm{D}_{\mbox{KL}}\left(q_{\phi}(\mathbf{z} |\mathbf{x}) \| p(\mathbf{z})\right) 
     - \gamma\sum_{m=1}^{M}\left(\delta  - \mathrm{H}(\mathbf{z}_m)\right).  \\
 \end{aligned}
\end{equation}
We summarize the difference in loss functions among vanilla VAE, 
$\beta$-VAE, and our approach in Table~\ref{table:LossFunc}.
Following the convention of Eq.~\eqref{eqn:VAElossfunc}, $\mathcal{L}_{V}$,
$\mathcal{L}_{\beta}$, and $\mathcal{L}_{\mathrm{H}}$ are all written as
objectives to be \emph{maximized}; the quantities minimized by the optimizer in
our implementation are their negatives.
\begin{table*}[htb]
\begin{center}
\caption{Loss Function Comparison}
\label{table:LossFunc}
\begin{tabular}{llll} 
\hline
Loss Functions                   & Accuracy                                                                    & KL Divergence Constraint                                            & Entropy-based Constraint   \\ 
\hline
$\mathcal{L}_{V}(\theta, \phi ; \mathbf{x},\hat{\mathbf{x}}, \mathbf{z})$ - Vanilla VAE& 
$\mathbb{E}_{q_{\phi}(\mathbf{z} | \mathbf{x})}\left[\ln p_{\theta}(\hat{\mathbf{x}} | \mathbf{z}))\right]$ &
$-\mathrm{D}_{\mbox{KL}}\left(q_{\phi}(\mathbf{z}|  \mathbf{x}) \| p(\mathbf{z})\right)$ &
\\ 
\hline
$\mathcal{L}_{\beta}(\theta, \phi, \beta; \mathbf{x},\hat{\mathbf{x}}, \mathbf{z})$ - $\beta$-VAE &
$\mathbb{E}_{q_{\phi}(\mathbf{z} | \mathbf{x})}\left[\ln p_{\theta}(\hat{\mathbf{x}} | \mathbf{z}))\right]$&
$-\beta\mathrm{D}_{\mbox{KL}}\left(q_{\phi}(\mathbf{z}|  \mathbf{x}) \| p(\mathbf{z})\right)$      &  
\\ 
\hline
$\mathcal{L}_{\mathrm{H}} (\theta, \phi, \beta, \gamma ; \mathbf{x},\hat{\mathbf{x}}, \mathbf{z}) $ - \bf{Ours} & 
$\mathbb{E}_{q_{\phi}(\mathbf{z} | \mathbf{x})}\left[\ln p_{\theta}(\hat{\mathbf{x}} | \mathbf{z}))\right]$ &
$-\beta\mathrm{D}_{\mbox{KL}}\left(q_{\phi}(\mathbf{z}|  \mathbf{x}) \| p(\mathbf{z})\right)$      &
$-\gamma\sum_{m=1}^{M}\left(\delta  - \mathrm{H}(\mathbf{z}_m)\right)$\\ 

\hline
\end{tabular}
\end{center}
\end{table*}

\subsection{Optimization of Latent-space Dimensionality}
\label{sec:Optlatent}
\begin{figure}[thb]
\centering
\includegraphics[width=0.45\textwidth]{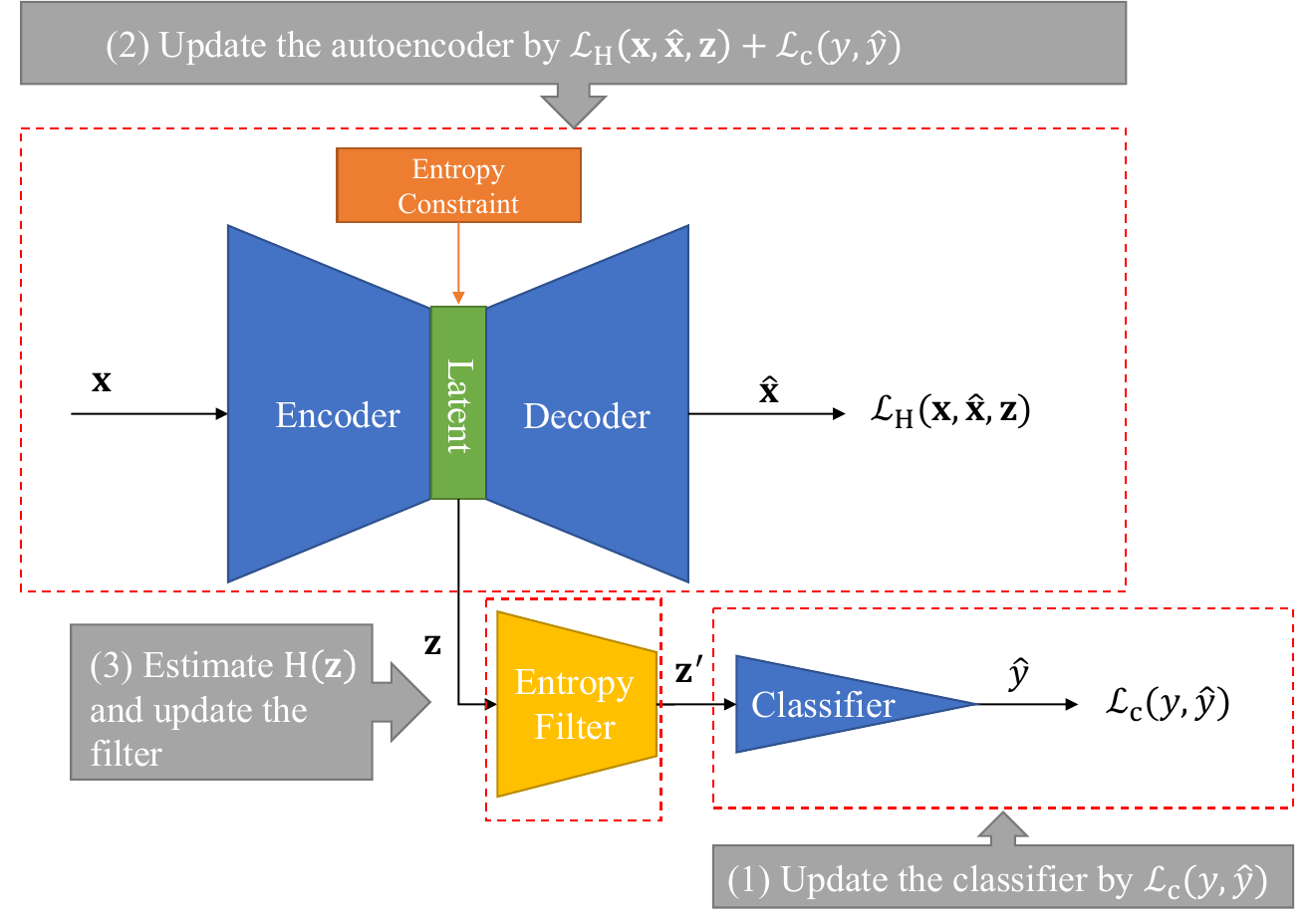}
\caption{A schematic of the optimization of latent-space dimensionality.
In each epoch, (1) the classifier is updated with $\mathcal{L}_\mathrm{C}$,
(2) the autoencoder is updated with $\mathcal{L}_{\text{AE}}$, and (3) the
latent-variable entropies are re-estimated and the entropy filter is updated.}
\label{fig:OnlineClassfier}
\end{figure}

We present a weight-filter method that leverages the soft-constraint property to
optimize the latent-space dimensionality, and thereby addresses the outer problem
of Eq.~\eqref{eqn:bilevel}.
The method is implemented in the context of a classification task.
The classifier and the autoencoder are updated in alternation within each epoch:
the classifier descends $\mathcal{L}_\mathrm{C}$ with the encoder held fixed, and
the autoencoder then descends $\mathcal{L}_{\text{AE}}$ of
Eq.~\eqref{eqn:AdvAELoss}, which contains $\mathcal{L}_\mathrm{C}$ as one of its
terms.
The classification loss therefore enters the autoencoder objective as a penalty:
it constrains the encoder to retain whatever the classifier needs, while the
reconstruction and entropy terms constrain it in their own directions.
The method builds on the EC discussed previously.
We introduce an entropy filter into the latent space, a linear filter whose
weights are determined by the entropies of the latent variables.
The filtered latent space $\mathbf{z}^\prime$ is obtained as the element-wise
product
\begin{equation}
\label{eqn:filter}
    \mathbf{z}^\prime = \mathbf{w} \odot \mathbf{z},
\end{equation}
where $\odot$ denotes the element-wise (Hadamard) product 
and $\mathbf{w}$ is a non-negative weight vector whose zero entries mark the pruned dimensions, 
$\mathbf{w} = {\begin{bmatrix}
w_1, w_2,\cdots,w_M
\end{bmatrix}}^\mathsf{T}$,
where the superscript $\mathsf{T}$ denotes transpose operation.
A zero weight removes the corresponding dimension from the representation seen by
the downstream task: it indicates that the dimension carries too little entropy
to be worth retaining, and that removing it does not violate the accuracy
constraint.
The filtered latent variables serve as the input to the classifier, whose loss
acts as an additional soft constraint on the autoencoder.
The optimization is formulated by maximizing the expected log-likelihood 
of the reconstruction while satisfying the EC 
and the classification loss constraint, 
 \begin{equation}
\begin{aligned}
 \label{eqn:AdversialTraining}
    \max_{\phi, \theta}\quad &{  \mathbb{E}_{q_{\phi}(\mathbf{z} | \mathbf{x})}\left[\ln p_{\theta}(\hat{\mathbf{x}} | \mathbf{z}))\right]} \quad \\[0.1in]
     \textrm{s.t.} \quad & \mathrm{D}_{\mbox{KL}}\left(q_{\phi}(\mathbf{z} | \mathbf{x}) \| p(\mathbf{z})\right)<\epsilon_\mathrm{K} \\
     &\sum_{m=1}^{M}\left(\delta-   \mathrm{H}(\mathbf{z}_m)\right) < \epsilon_\mathrm{H}\\
     & \mathcal{L}_\mathrm{C}\left(y,\hat{y}\right) < \epsilon_\mathrm{C},
\end{aligned}
\end{equation}
where $\mathcal{L}_\mathrm{C}\left(y,\hat{y}\right)$ 
is the loss function of the classifier, 
$y$ is the ground-truth label, and $\hat{y}$ is the predicted label. 
The classifier loss $\mathcal{L}_\mathrm{C}$ is a cross-entropy loss over the ten
digit classes, computed from the sigmoid outputs of the classifier described in
Section~\ref{sec:experiment}.
Combining the negated autoencoder objective with the classifier loss gives the
total loss that is minimized with respect to $\phi$ and $\theta$,
\begin{equation}
\begin{aligned}
\label{eqn:AdvAELoss}
    \mathcal{L}_{\text{AE}} (\mathbf{x},\hat{\mathbf{x}}, \mathbf{z}, y,\hat{y}) = &-\mathcal{L}_{\mathrm{H}} (\mathbf{x},\hat{\mathbf{x}}, \mathbf{z}) +  \mathcal{L}_\mathrm{C}\left(y,\hat{y}\right),\\
 \end{aligned} 
\end{equation}
The proposed optimization method of latent-space dimensionality is illustrated in Fig.~\ref{fig:OnlineClassfier}.

The weights of the entropy filter are updated using Algorithm~\ref{alg:FilterWeight}.
A threshold $\alpha$ of the classifier accuracy controls the reduction of the dimensionality.
The entropy filter first sorts the latent variables in descending order of
entropy.
If the accuracy of the classifier is above the threshold $\alpha$, the filter
eliminates the lowest-entropy latent variable that is still being fed to the
classifier; otherwise the weights are left unchanged and the current
dimensionality is retained.
The retained weights are set to $1/J$ rather than to unity, so that the total
scale of the filtered code $\mathbf{z}^\prime$ is preserved as $J$ decreases and
the classifier is not presented with an input whose magnitude grows as dimensions
are removed.
One consequence should be noted explicitly: because the scale factor changes
whenever a dimension is dropped, the classifier sees a rescaled input at each
reduction step, and the convergence comparison of Fig.~\ref{fig:optdim} measures
the combined effect of the reduced dimensionality and this rescaling.
Both models in that comparison are subject to the same rescaling, so it does not
bias the comparison between them; we return to the point in
Section~\ref{sec:limitations}.
The output of the algorithm is the updated weight array and the resulting
dimensionality. 

\begin{algorithm}[htbp]
  \caption{Entropy-Filter Weight Update Algorithm}
  \label{alg:FilterWeight}
 \begin{algorithmic}[1]
\renewcommand{\algorithmicrequire}{\textbf{Input:}}
 \renewcommand{\algorithmicensure}{\textbf{Output:}}
\REQUIRE $\mathbf{H}$ - entropy array of the latent variables, \\$\mathbf{H}=\{H_1, H_2, \cdots, H_M\}$,\\
$\mathbf{w}$~-~current filter weight array, \\
$a$~-~current classifier training accuracy, \\
$\alpha$~-~training accuracy threshold.\\
 \ENSURE $\mathbf{w}$ - updated weight array of the entropy filter,\\
 $J$~-~number of retained latent dimensions.\\
 \textbf{Initialization:}
 \STATE Obtain the number of currently retained dimensions,
 \\ $J\leftarrow\text{sum}(\mathbf{w}>0)$.\\
 \textbf{Update:}
 \STATE Sort $\mathbf{H}$ in descending order and obtain the index array $j = \{j_1, j_2,\cdots,j_M\}$, so that $j_1$ is the index of the latent variable with the highest entropy and $j_M$ that with the lowest.
\IF{$a>\alpha$ \textbf{and} $J>1$}
\STATE Set $J \leftarrow J-1$ to remove one dimension.
 \STATE Retain the $J$ highest-entropy latent variables and eliminate the rest by setting their weights to $0$,
 $$
 w_{j_i} =
 \begin{cases}
\frac{1}{J},\quad &i\leq J\\
  0, &i> J
 \end{cases}.
 $$
 \ENDIF
 \RETURN $\mathbf{w}$, $J$
 \end{algorithmic}
 \end{algorithm}

Estimating the entropy in the training process poses a challenge 
due to the lack of a well-established distribution of the latent space. 
To address this challenge, we propose an entropy-estimation algorithm 
that utilizes batch training as outlined in Algorithm~\ref{alg:EstEntropy}.
The algorithm computes a histogram of each latent variable for every batch in an
epoch and accumulates the bin counts over the epoch.
The accumulated counts are normalized to approximate the probability mass function
(PMF) of the binned variable, and the entropy is estimated by the plug-in formula
applied to that PMF.
Two details matter for reproducibility.
First, the bin edges must be fixed across batches and epochs, since otherwise the
estimates are not comparable over training.
Second, the resulting quantity is the discrete entropy of the binned variable,
which exceeds the differential entropy of Eq.~\eqref{eqn:infoEtropy} by
approximately $\log(1/\Delta)$ for bins of width $\Delta$; the offset is constant
across models and therefore does not affect the comparisons we report, but it does
mean that the numerical values are not on the $\frac{1}{2}\ln(2\pi e)$ scale used
to set $\delta$.
\begin{algorithm}[htbp]
    \caption{Entropy Estimation Algorithm}
    \label{alg:EstEntropy}
    \begin{algorithmic}[1]
     \renewcommand{\algorithmicrequire}{\textbf{Input:}}
 \renewcommand{\algorithmicensure}{\textbf{Output:}}
 \REQUIRE $\mathrm{z}_m$ - a latent variable batch, $m\in[1,M]$ , \\
 $M$ - number of latent variables, \\
 $N$ - number of samples in the training set, \\
 $K$ - number of bins in the histogram, spanning a fixed range shared by all batches.
 \ENSURE  $H_m$ - estimated entropy of the latent variable $\mathbf{z}_m$.
 \STATE Initialize the bin counts, $\mathbf{B}\leftarrow\mathbf{0}$.
 \FOR{each batch in an epoch}
 \STATE Compute the histogram of $\mathrm{z}_m$ over the $K$ fixed bins and store the counts in an array $\mathbf{b} = \{b_1, b_2, \cdots, b_K\}$.
 \STATE Accumulate the counts, $\mathbf{B} \leftarrow \mathbf{B}+\mathbf{b}$, $\mathbf{B} = \{B_1, B_2, \cdots, B_K\}$.
 \ENDFOR
 \STATE Normalize the total bin counts by $\mathbf{B}\leftarrow \frac{\mathbf{B}}{N}$, which approximates the PMF of the binned latent variable.
\STATE Estimate the entropy as $H_m \leftarrow -\sum_{k=1}^{K} B_k \log{B_k}$, with the convention $0\log 0 = 0$.
\RETURN $H_m$
    \end{algorithmic}
\end{algorithm}

The training procedure for optimizing the dimensionality of latent space is described in Algorithm~\ref{alg:training}. 
The algorithm initializes the entropy-filter weights uniformly. 
Within each epoch, the classifier and the autoencoder parameters are updated over
each batch by minimizing $\mathcal{L}_\mathrm{C}$ and $\mathcal{L}_{\text{AE}}$,
respectively.
The classifier accuracy $a$ is then recorded and each latent-variable entropy is
estimated using Algorithm~\ref{alg:EstEntropy}. 
Finally, the entropy-filter weights are updated 
using Algorithm~\ref{alg:FilterWeight}, 
which determines the optimal dimensionality of the latent space $J$ 
by identifying the subset of latent variables with high entropy values. 
After the classifier and the autoencoder have converged, 
the latent variables $\mathbf{z}^\prime$ associated with the strictly positive
entries of $\mathbf{w}$ constitute the retained latent space. 
The output of Algorithm~\ref{alg:training} is the resulting dimensionality $J$ 
and the corresponding weight vector $\mathbf{w}$.
Because the dimensionality is only ever decremented, and only while the accuracy
threshold holds, the procedure returns a feasible value of $J$ rather than a
certified minimum; a dimension eliminated early is never reinstated.
\begin{algorithm}[htb]
\caption{Training Procedure for Optimization of Latent-space Dimensionality }
\label{alg:training}
\begin{algorithmic}[1]
\STATE Initialize each entropy filter weight evenly by $w_m = \frac{1}{M}$.
\FOR{each epoch}
\FOR{each batch}
\STATE Update the classifier parameters using the loss function $\mathcal{L}_\mathrm{C}$.
\STATE Update the autoencoder parameters by minimizing the total loss $\mathcal{L}_{\text{AE}}$ in Eq.~\eqref{eqn:AdvAELoss}.
\ENDFOR
\STATE Obtain the classifier accuracy $a$.
\STATE Estimate the latent-variable entropies by Algorithm~\ref{alg:EstEntropy}.
\STATE Update the entropy filter weights by Algorithm~\ref{alg:FilterWeight} and obtain the new dimensionality of latent space $J$.
\ENDFOR
\STATE Return $J$ as the resulting dimensionality; the retained latent space
corresponds to the strictly positive entries of $\mathbf{w}$.
\end{algorithmic}
\end{algorithm}

The weight filter was designed to exploit the slack of the EC: the dimensions it
removes are those for which the constraint of
Eq.~\eqref{eqn:EntropyConstraint0} is satisfied least well.
It does not, however, depend on the EC being present.
Over-pruning by itself depresses the entropy of the affected latent variables, so
the ranking the filter needs is available even when $\gamma=0$, and the filter can
be applied to a plain VAE.
Sections~\ref{sec:experiment} and~\ref{sec:discussion} report both cases.

\section{Experimental Results}\label{sec:experiment}
Experiments were performed to verify and validate the improvement 
of EC of latent variables 
and the optimization of latent-space dimensionality. 
We first examined the latent-variable encoding capacity 
by imposing the EC. 
Then we evaluated the performance of the optimization of latent-space dimensionality in a classification task.

\subsection{Setup}
\textbf{Datasets:}
We employed two datasets in our experiments. 
The first is dSprites~\cite{dsprites17}, a synthetic dataset introduced by $\beta$-VAE, 
comprising over 700,000 images of 2D shapes with five different ground-truth labels (Table~\ref{table:dSpritesLabels}).
This dataset serves to assess the disentanglement and encoding capacity of latent variables. 
The second dataset is MNIST~\cite{Deng_2012}, 
a collection of handwritten digits containing 60,000 training samples and 10,000 testing samples,
each represented as a grayscale $28\times28$ image.
We utilized the dSprites dataset for conducting experiments on the EC of latent variables. 
This dataset was chosen due to its labeled data features, which allowed us to investigate the impact of the EC on specific attributes. 
For the purpose of the optimization of latent-space dimensionality, 
we employed the MNIST dataset. 
MNIST is widely used in such optimization tasks as it consists of real-world data, enabling us to evaluate the performance of our approach on practical image data.
\begin{table}[htbp!]
    \caption{dSprites Dataset Labels\label{table:dSpritesLabels}}
    \centering
\begin{tabular}{lll}
\hline
Label & Range & Classes\\ \hline
Shape & 1,2,3 (square, ellipse, heart) & 3\\
Scale& [0.5, 1] & 6 \\
Orientation & [0, $2\pi$] & 40 \\
$X$-Position &[0, 1] &32 \\
$Y$-Position &[0, 1] &32 \\
\hline
\end{tabular}
\end{table}

\textbf{VAE Reference Models: }
Table~\ref{table:modelsettings} lists the autoencoder configurations used for each dataset.
We used the same latent-space dimension, $M=10$, 
as the reference models of $\beta$-VAE~\cite{higgins2017beta},~\cite{burgess2018understanding}.
For the MNIST dataset, we used the multilayer perceptron (MLP) model identical to the vanilla VAE's setup.
For the downstream task we built an MLP classifier with 100 hidden nodes, 
a ReLU activation at the hidden layer, 
and a sigmoid activation at the output layer.

\begin{table}[htbp!]
    \caption{Experimental Autoencoder Settings}
    \label{table:modelsettings}
\begin{tabular}{p{0.6cm}p{0.7cm}lp{3cm}lp{0.5cm}lll}
\hline
Dataset & Optimizer &              & Architecture                                        & Activation \\ \hline

dSprites & Adam      & Input        & $64\times 64$                                       &            \\
         & learning rate 5e-4& Encoder      & Conv $32\times 4\times 4$ (stride 2)                & ReLU           \\
         &  &              & Conv $32\times 4\times 4$ (stride 2)                & ReLU           \\
         &     &              & Conv $32\times 4\times 4$ (stride 2)                & ReLU           \\
         &           &              & Conv $32\times 4\times 4$ (stride 2)                & ReLU           \\
         & batch size: 128           &              & FC 784,400        & ReLU       \\
         &           & Latent space & 10                                             &            \\
         & Epochs: 200          & Decoder      & Deconv reverse of encoder  & ReLU       \\
         &           & Output       & $64\times 64$                                       & Sigmoid\\
         \hline
         MNIST    & Adam      & Input        & $28\times 28$                                       &            \\
         & learning rate 1e-2      & Encoder      & FC 784,400        & ReLU       \\
         &  batch size: 200           & Latent space & 10                                             &            \\
         & Epochs: 300          & Decoder      & FC 400,784                                          & Sigmoid     \\
         &           & Output       & $28\times 28$                                       &            \\ \hline
\end{tabular}
\end{table}

\textbf{Loss Function: }
The loss function in Eq.~\eqref{eqn:newproblemoriginloss} 
consists of three terms: 
the reconstruction loss, KL divergence, and the EC.
The KKT multipliers $\beta$ and $\gamma$ control the strength of the KL divergence term and the EC, respectively. 
A lower bound for the entropy of each latent variable is set by $\delta$.
When $\beta=1$, the autoencoder is equivalent to the vanilla VAE.
We used $\beta = 4$ to match the $\beta$-VAE~\cite{burgess2018understanding} for comparison purposes.
We chose $\gamma = 1$ and $\delta=1$ to qualitatively evaluate the effect of the EC.

\textbf{Training: } 
For reproducibility, we fixed the random seed for the reparameterization trick in
all experimental models.
The multipliers $\beta=4$, $\gamma=1$, and the target $\delta=1$ were set a priori
rather than tuned, so that the effect of the EC could be read off directly.
The single exception is the EC$^*$ variant: for the $\beta$-VAE family we ran a
10-trial Bayesian search with Optuna~\cite{optuna_2019} over $\gamma$, minimizing
the reconstruction loss, which returned $\gamma=0.36$.
The EC$^*$ models therefore indicate the reconstruction quality attainable when
$\gamma$ is tuned, while all other models are untuned; this asymmetry should be
kept in mind when reading Tables~\ref{table:beta4results}
and~\ref{tbl:betametrics}.

We use the following abbreviations for our methods in the models:
\begin{itemize}
\item
EC: Entropy-based Constraint - Imposes the EC on latent variables with $\gamma = 1$.
\item
LB: Lower-Bound - Sets a lower-bound on the entropy of latent variables, ensuring a minimum information content within the latent variable. We set the lower-bound $\delta = 1$.
\item
EC*: KKT Optimization - Involves fine-tuned $\gamma=0.36$, the KKT multiplier of the EC.
\end{itemize}

\subsection{Evaluation Metrics}
We evaluate the performance of the latent space by considering both latent variable capacity and disentanglement. 
To achieve this, we utilize four distinct high-level metrics.

\subsubsection{Latent Variable Activation Score (LVAS)} 
LVAS was initially introduced in IWAE~\cite{burdaImportanceWeightedAutoencoders2016}. 
This score assesses the encoding capacity of a latent variable. 
If a latent variable effectively captures information from the data, its distribution is expected to vary with different observations. 
Mathematically, LVAS is defined as the activity statistics of a latent variable:
\begin{equation}
\label{eqn:LVAS}
    A_{\mathbf{z}_m} = \mbox{Cov}_\mathbf{x}\left(\mathbb{E}_{\mathbf{z}_m\sim q(\mathbf{z}_m|\mathbf{x})}\left[\mathbf{z}_m\right]\right),
\end{equation}
which, for a scalar latent variable, is the variance across inputs of the
posterior mean.

A higher LVAS indicates superior encoding capacity.
Additionally, we define a latent variable as "active" when $A_{\mathbf{z}_m} > 0.01$, 
signifying that the latent variable is not pruned. 
Generally, a greater number of active latent variables (NALV) tends to result in improved downstream representations~\cite{sinhaConsistencyRegularizationVariational2022}, although exceptions exist.

A count of dimensions above a fixed threshold is a coarse statistic, however,
since it weighs a marginally active dimension the same as a strongly active one.
To evaluate the encoding capacity of the latent space as a whole we therefore also
report the total LVAS summed over the active latent variables, and we base our
comparisons on that total rather than on the count.

\subsubsection{\texorpdfstring{$\beta$-VAE Metric}{Beta-VAE Metric}}\label{sec:disentanglement}
The $\beta$-VAE metric~\cite{higgins2017beta}
assesses the disentanglement of the entire latent space.
This metric employs the variance of the latent vector to train a linear classifier,
which predicts the corresponding labeled feature. 
The accuracy of this classifier serves as an index for evaluating disentanglement performance.

\subsubsection{FactorVAE  Metric}
The FactorVAE metric~\cite{Kim_Mnih} complements the $\beta$-VAE metric
by normalizing the latent vector as input to a linear model, 
thereby expanding the quantitative range of the metric score beyond the variance of the latent vector.
A voter classifier is then employed to tally the frequency with which an individual latent variable dominates an individual labeled feature. 
The accuracy score is derived from the proportion of this frequency returned by the voter classifier.

\begin{figure}[htb]
\centering
\includegraphics[width=0.45\textwidth]{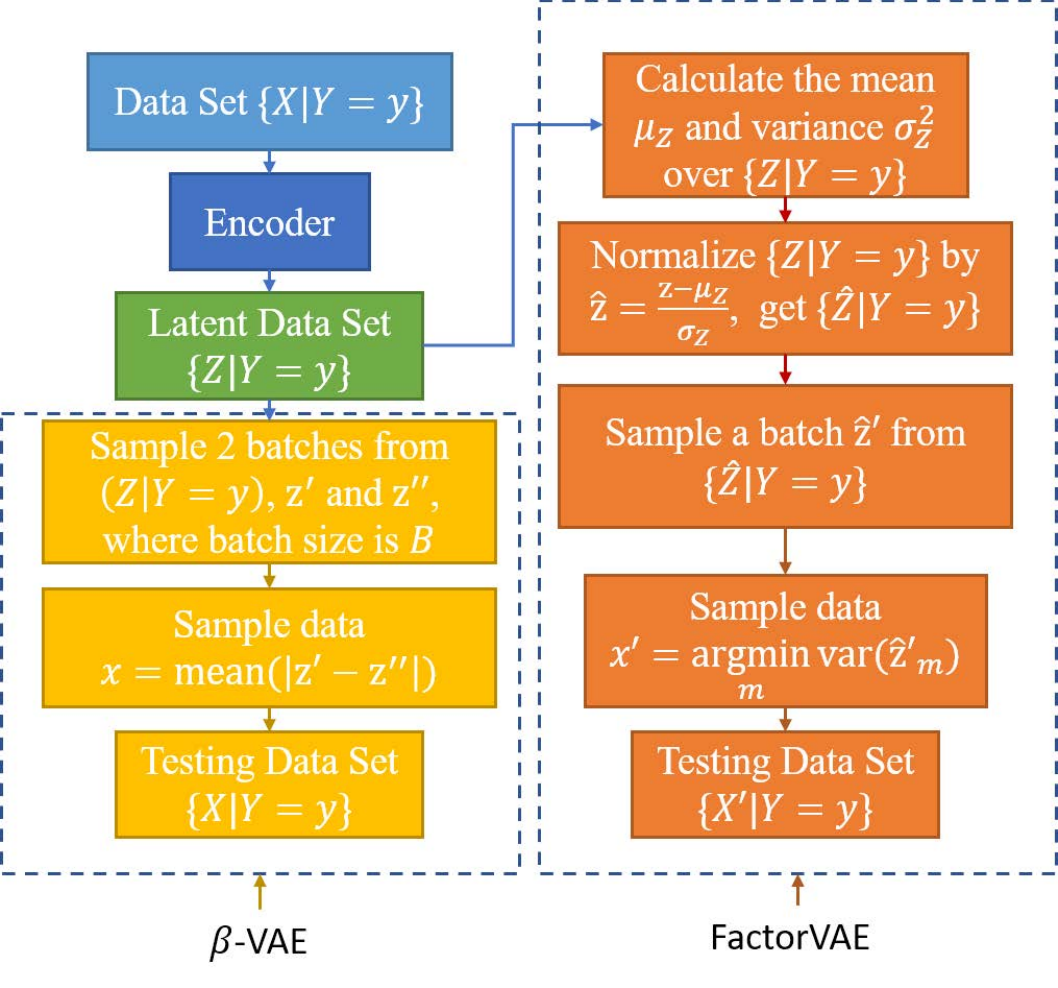}
\caption{Comparison of data preprocessing for $\beta$-VAE metric and FactorVAE metric. 
The left flow chart shows the data preparation process for the $\beta$-VAE metric, while the right flow chart shows the process for the FactorVAE metric.}
\label{fig:DisentanglementMetrics}
\end{figure}

Both $\beta$-VAE metric and Factor VAE metric use classifier accuracy as a measure of disentanglement, 
and a higher score indicates a better disentanglement. 
However, the $\beta$-VAE metric is more sensitive 
to the overall information preserved in the latent space, 
while the FactorVAE metric is more sensitive 
to the independent relationship between an individual latent variable and a labeled feature. 
These differences in sensitivity are due to the different preprocessing 
of the data as shown in Fig.~\ref{fig:DisentanglementMetrics}.

\subsubsection{Correlation-based Metric}
In our implementation, we have observed instances where certain labels are occasionally encoded in more than one latent variable. 
This occurrence may be attributed to either a labeled feature surpassing the encoding capacity of a latent variable 
or a lack of well-disentangled features leading to mixed encoding.

To address this issue, we propose a correlation-based metric capable of compensating for such occurrences. 
For each labeled feature, the metric takes the two largest absolute correlations
between that feature and any latent variable, and sums them over all features, as
outlined in Algorithm~\ref{alg:corrScore}.
The score therefore lies in $[0,2L]$, where $L$ is the number of labeled features;
with the $L=5$ features of dSprites the maximum is $10$, and dividing by $2L$
gives a normalized score in $[0,1]$.
\begin{algorithm}[htbp]
\caption{Correlation Score Algorithm}
\label{alg:corrScore}
    \begin{algorithmic}[1]
         \renewcommand{\algorithmicrequire}{\textbf{Input:}}
 \renewcommand{\algorithmicensure}{\textbf{Output:}}
    \REQUIRE $\{\mathbf{z}, \mathbf{l}\}$ - a set of all latent variables and all label values, $\mathbf{z}= \{\mathbf{z}_1,\mathbf{z}_2,\cdots, \mathbf{z}_M\}$, $\mathbf{l}= \{\mathbf{l}_1,\mathbf{l}_2,\cdots, \mathbf{l}_L\}$, where $M$ is the
total number of latent variables and $L$ is the total number
of labels.
    \ENSURE $S$ - the correlation score
        \STATE Compute the correlation absolute values, $$c = |\mbox{Corr}(\mathbf{z},\mathbf{l})|,\quad c\in\mathcal{R}^{M\times L}   
        $$
        \STATE Sort each column of $c$ by descending order as
        $$c_{l} = \begin{Bmatrix} 
        c_{l,(1)},c_{l,(2)},\cdots,c_{l, (M)}
        \end{Bmatrix}^\top,\quad l\in[1,L]$$
        \STATE Compute the score by the top two correlations with each label, $$S =\sum_{l=1}^{L} \sum_{i=1}^{2}c_{l,(i)}$$
    \end{algorithmic}
\end{algorithm}

Our metric draws inspiration from Canonical Correlation Analysis (CCA),
which addresses the correspondence between two data sets.
Specifically, our approach is akin to a n-component CCA,
providing insights into both encoding capacity through correlation 
and disentanglement by limiting the component number to 2.

\subsection{Statistics of the Latent-variable Entropy}
We applied both the vanilla VAE and $\beta$-VAE to encode the dSprites dataset. 
Experimental statistics of latent-variable entropy of the vanilla VAE models are presented in Table~\ref{table:vanillaresults},
while those of the $\beta$-VAE models are detailed in Table~\ref{table:beta4results}. 
The reconstruction error, representing the average cross-entropy between the input and output,
is also provided in the tables.
The entropies are the plug-in estimates returned by Algorithm~\ref{alg:EstEntropy}
and are therefore discrete entropies of the binned latent variables, on the scale
discussed in Section~\ref{sec:Optlatent}; they are comparable across the rows of a
table but not directly against the differential-entropy target $\delta$.

Every EC variant raised the average latent-variable entropy above its own
baseline: from $5.1163$ to $5.3573$ and $5.2349$ in the vanilla family, and from
$5.1432$ to between $5.1641$ and $5.2409$ in the $\beta$-VAE family.
The increase is not uniform across the three statistics, however.
The maximum entropy of VAE+EC+LB ($5.3474$) and the minimum entropy of
$\beta$-VAE+EC*+LB ($4.9239$) both fall below the corresponding baseline values,
so the EC raises the entropy of the latent space in aggregate rather than of every
latent variable individually.
Reconstruction error behaves less uniformly still.
VAE+EC attained the lowest reconstruction error of the entire study
($19.4627$, a 38\% reduction relative to the vanilla VAE at $31.2481$), and
$\beta$-VAE+EC*+LB attained the lowest error among the $\beta$-VAE models
($28.4817$ against $29.6556$, a 4\% reduction).
The remaining variants were worse than their baselines, with $\beta$-VAE+EC+LB
the worst at $46.1787$.
The EC can therefore relieve the reconstruction penalty associated with
over-pruning, but the size of $\gamma$ and the presence of the lower bound matter,
and a poor choice makes reconstruction worse rather than better.
The spread across the five $\beta$-VAE rows, from $28.48$ to $46.18$, is wide
enough that we do not claim the ordering within it is significant on the basis of
a single seed; Section~\ref{sec:limitations} states what would be required to
establish that.

\begin{table}[htbp]
\caption{Vanilla VAE Latent-variable Entropy and Reconstruction Error.
Entropies are the plug-in estimates of Algorithm~\ref{alg:EstEntropy} and are
comparable across rows but not against the differential-entropy target $\delta$.
Bold marks the best value in each column.}
\label{table:vanillaresults}
\begin{tabular}{lcccc}
\hline
\multicolumn{1}{c}{}  & \multicolumn{3}{c}{Latent Variable Entropy}                & \multicolumn{1}{c}{} \\ \hline
Model                & Ave.               & Min.     & Max.              & Reconstruction Error \\ \hline
Vanilla VAE          & 5.1163          & 4.9495          & \textbf{5.4181} & 31.2481              \\
VAE+EC             & \textbf{5.3573} & \textbf{5.0792} & 5.4167          & \textbf{19.4627}     \\
VAE+EC+LB             & 5.2349          & 5.0038          & 5.3474          & 32.4894              \\
\hline
\end{tabular}
\end{table}

\begin{table}[htbp]
\centering
\caption{$\beta$-VAE Latent-variable Entropy and Reconstruction Error.
Entropies are on the same scale as Table~\ref{table:vanillaresults}.
Bold marks the best value in each column.}
\label{table:beta4results}
\begin{tabular}{lcccc}
\hline
\multicolumn{1}{c}{}  & \multicolumn{3}{c}{Latent Variable Entropy}                  & \multicolumn{1}{c}{Loss} \\ \hline
Model                & Ave.            & Min.            & Max.            & Reconstruction Error \\ \hline
$\beta$-VAE          & 5.1432          & 4.9593          & 5.2201          & 29.6556              \\
$\beta$-VAE+EC*            & 5.1743          & 5.0122          & 5.2302          & 40.3712              \\
$\beta$-VAE+EC*+LB             & 5.1641          & 4.9239          & 5.2174          & \textbf{28.4817}     \\
$\beta$-VAE+EC             & 5.2386          & 5.0959          & 5.2796          & 35.2022              \\
$\beta$-VAE+EC+LB             & \textbf{5.2409} & \textbf{5.1566} & \textbf{5.2865} & 46.1787   \\  
\hline
\end{tabular}
\end{table}

\subsection{Analysis of Encoding Capacity and Disentanglement of Latent Variables}
We conducted a comparison between the vanilla VAE and $\beta$-VAE using the evaluation metrics presented in Table~\ref{tbl:vanillametrics} and Table~\ref{tbl:betametrics}. 
The disentanglement scores, derived from both the $\beta$-VAE metric and the FactorVAE metric, 
represent the average scores across all labels in the dSprites dataset.

The two families behave differently, so we state the comparison per family rather
than in aggregate.

In the vanilla-VAE family (Table~\ref{tbl:vanillametrics}), the EC increased the
total LVAS from $7.13$ to $10.18$ and $11.58$---gains of 43\% and 62\%---and the
correlation metric from $3.6074$ to $3.9361$ and $4.1103$, gains of 9\% and 14\%.
Both disentanglement metrics, however, fell slightly below the vanilla VAE: the
$\beta$-VAE metric from $0.5733$ to $0.5393$ and $0.5272$, and the FactorVAE
metric from $0.7454$ to $0.7118$ and $0.7115$.

In the $\beta$-VAE family (Table~\ref{tbl:betametrics}), $\beta$-VAE+EC attained
the highest total LVAS in this family ($6.11$ against $5.23$, a gain of 17\%),
the highest FactorVAE score of any model in the study ($0.8907$, against $0.8468$
for the $\beta$-VAE baseline), and the highest correlation score of any model
($4.1268$).
The $\beta$-VAE metric moved in the opposite direction for three of the four
variants; only the fine-tuned $\beta$-VAE+EC*+LB ($0.5118$) exceeded the
$\beta$-VAE baseline ($0.5079$), and $\beta$-VAE+EC+LB fell to $0.4172$.

Two conclusions follow.
The EC consistently increases the activation-based and correlation-based measures
of capacity, which is what it was designed to do.
Its effect on the two classifier-based disentanglement metrics is regime
dependent: it improves the FactorVAE metric under strong KL regularization
($\beta=4$), where that metric rewards the one-to-one label--variable
correspondence our constraint encourages, and it does not improve the $\beta$-VAE
metric in either regime.
Since the $\beta$-VAE metric is the more sensitive of the two to the total
information retained in the latent space (Section~\ref{sec:disentanglement}), and
since the EC redistributes rather than uniformly increases that information, this
asymmetry is consistent with the mechanism rather than evidence against it.
We are careful not to overstate this reading.
Two metrics that disagree cannot by themselves establish which of them is
tracking the property of interest, and the disagreement here may reflect the
metrics as much as the representation.
A third, differently constructed measure would settle it, and
Section~\ref{sec:limitations} identifies this as the principal gap in the present
evidence.

\begin{table*}[htbp]
\caption{The Capacity of Latent Variables in Vanilla VAEs.
Bold marks the best value in each column.
The sixth active dimension counted for the vanilla VAE is borderline
(LVAS $=0.15$ against a threshold of $0.01$); see Fig.~\ref{fig:LVAS}a.}
\label{tbl:vanillametrics}
\centering
\begin{tabular}{lllccc}
\hline
\multicolumn{1}{c}{}   & \multicolumn{2}{c}{Activation}&\multicolumn{2}{c}{Disentanglement}    &           Correlation\\ \hline
Model                &  NALV &Total LVAS&$\beta$-VAE Metric & FactorVAE Metric & Corr. Metric      \\ \hline
Vanilla VAE          & \textbf{6}&7.13&\textbf{0.5733}    & \textbf{0.7454}  & 3.6074          \\
VAE+EC             & 5&10.18&0.5272             & 0.7118           & 3.9361          \\
VAE+EC+LB             & 5&\textbf{11.58}&0.5393             & 0.7115           & \textbf{4.1103} \\
\hline
\end{tabular}
\end{table*}

\begin{table*}[htbp]
\centering
\caption{The Capacity of Latent Variables in $\beta$-VAEs.
Bold marks the best value in each column.
EC$^*$ variants use $\gamma$ tuned by Optuna; all other rows use the untuned
$\gamma=1$.}
\label{tbl:betametrics}
\begin{tabular}{lllccc}
\hline
\multicolumn{1}{c}{}     &   \multicolumn{2}{c}{Activation}&\multicolumn{2}{c}{Disentanglement}       & Correlation\\ \hline
Model                &   NALV & Total LVAS&$\beta$-VAE Metric & FactorVAE Metric & Corr. Metric      \\ \hline
$\beta$-VAE          &   5& 5.23&0.5079             & 0.8468           & 4.0011          \\
$\beta$-VAE+EC*            &   5& 5.07&0.4745             & 0.8682           & 3.6699          \\
$\beta$-VAE+EC*+LB             &   \textbf{6}& 6.01&\textbf{0.5118}    & 0.8449           & 3.8894          \\
$\beta$-VAE+EC             &   5& \textbf{6.11}&0.4858             & \textbf{0.8907}  & \textbf{4.1268} \\
$\beta$-VAE+EC+LB             &   5& 5.53&0.4172             & 0.8801           & 3.7112          \\  
\hline
\end{tabular}
\end{table*}

\begin{figure}[htb] 
\centering
\subfloat[Vanilla VAEs]
{
\includegraphics[width=0.9\linewidth]{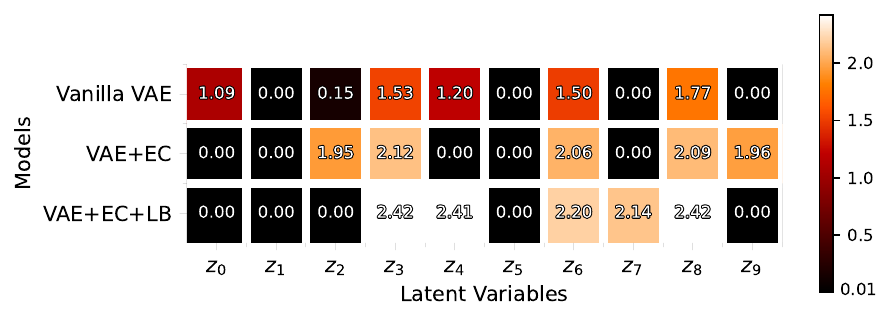}
}
\\
\centering
\subfloat[$\beta$-VAEs]
{
\includegraphics[width=0.9\linewidth]{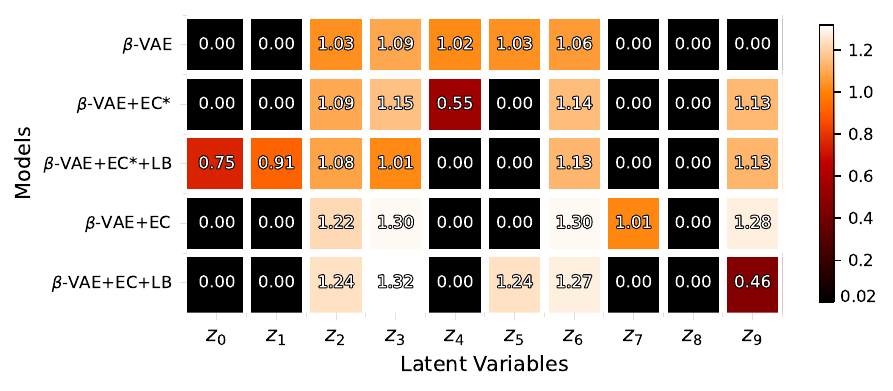}
}
\caption{Individual latent variable activation scores. 
The upper figure correspond to the vanilla VAE models,
while the lower figures pertain to the $\beta$-VAE models. 
The LVAS values are represented within each colored box, 
with the color indicating the range of values according to the accompanying color bars.
}
\label{fig:LVAS}
\end{figure}

The number of active latent variables (NALV) requires comment, because it is the
one column in which the vanilla VAE leads.
Figure~\ref{fig:LVAS} displays the individual scores.
The sixth active dimension of the vanilla VAE is $\mathbf{z}_3$, whose score of
$0.15$ only barely exceeds the activity threshold of $0.01$, whereas the five
active dimensions of VAE+EC and VAE+EC+LB carry substantially more activity each,
which is what produces their much higher totals.
A count of dimensions above a fixed threshold is thus a coarse summary: it treats
a marginally active dimension and a strongly active one alike.
We therefore report the total LVAS alongside the count and base our claims on the
former.
For the same reason we do not claim that the EC increases the number of active
latent variables; in the $\beta$-VAE family it does so for
$\beta$-VAE+EC*+LB (six against five) and in the vanilla family it does not.

\begin{figure}[htb] 
\centering
\subfloat[][$\beta$-VAE Metric\label{fig:vanillabetaDM}]{
\includepdftrim{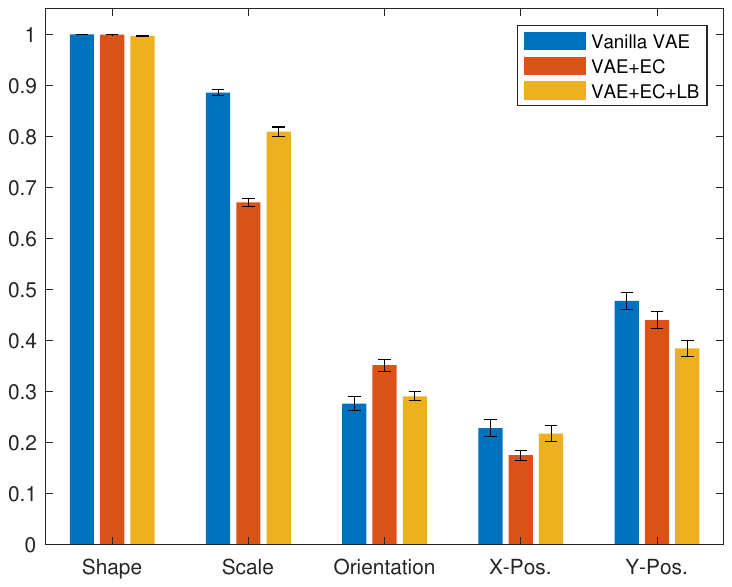}}
\hfill
\centering
\subfloat[][FactorVAE Metric\label{fig:vanillafactorDM}]{
\includepdftrim{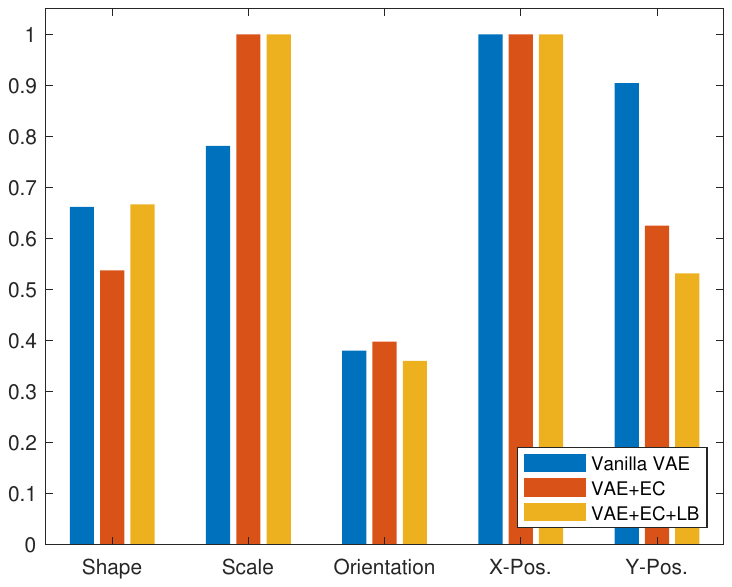}}
\caption{Vanilla VAE experiment disentanglement scores. 
The histograms show the disentanglement score of each label.}
\label{fig:vanillaDM}
\end{figure}

\begin{figure}[htb] 
\centering
\subfloat[$\beta$-VAE Metric\label{fig:betabetaDM}]{
\includepdftrim{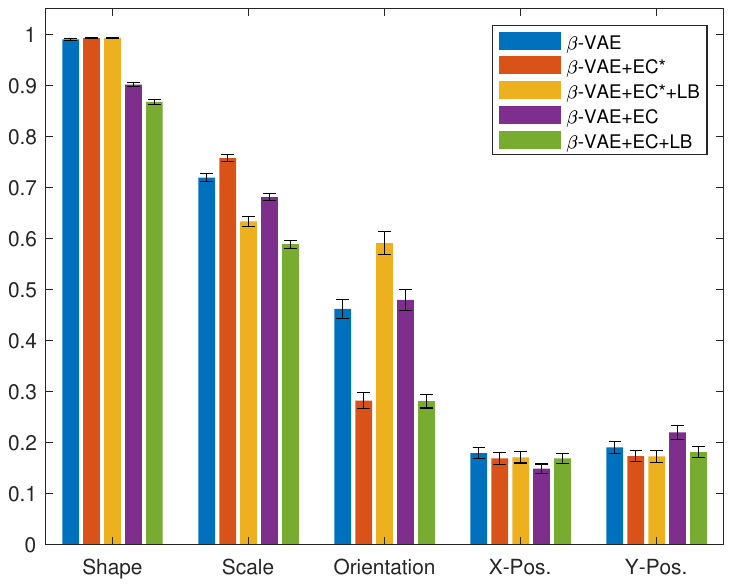}}
\hfill
\centering
\subfloat[FactorVAE Metric\label{fig:betafactorDM}]{
\includepdftrim{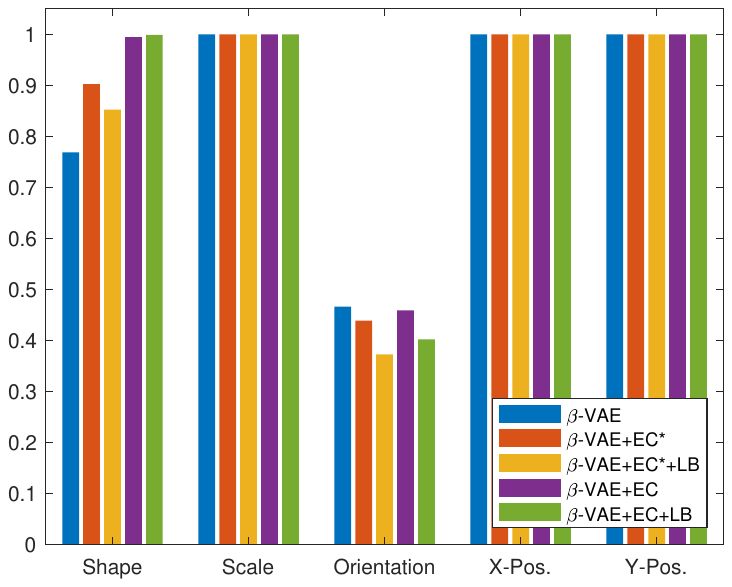}}
\caption{$\beta$-VAE experiment disentanglement scores. 
The histograms show the disentanglement score of each label.}
\label{fig:betaDM}
\end{figure}

Individual labeled-feature disentanglement scores are given 
in Fig.~\ref{fig:vanillaDM} and Fig.~\ref{fig:betaDM}.
When analyzing the detailed label $\beta$-VAE disentanglement scores in Fig.~\ref{fig:vanillabetaDM} from the vanilla VAE models,
our proposed approach outperformed the vanilla VAE model 
in encoding the orientation label, 
which was considered the most challenging to disentanglement~\cite{higgins2017beta,burgess2018understanding}.
In addition, our proposed approach achieved the highest FactorVAE disentanglement score, among the vanilla VAE and the $\beta$-VAE models, in encoding the scale label as shown in Fig.~\ref{fig:vanillafactorDM}.

Our fine-tuned model $\beta$-VAE+EC*+LB
achieved a significant improvement in the orientation label 
as shown in Fig.~\ref{fig:betabetaDM}. 
Furthermore, the FactorVAE score of the shape label, a discrete label feature, is higher in all of our models in Fig.~\ref{fig:betafactorDM} than in the $\beta$-VAE model. 
All of our models encoded the shape label in a more disentangled manner. 

\begin{figure}[htb] 
\centering
\subfloat[Vanilla VAE\label{fig:corr100}]
{
\includegraphics[width=0.5\linewidth, trim=40 0 40 0]{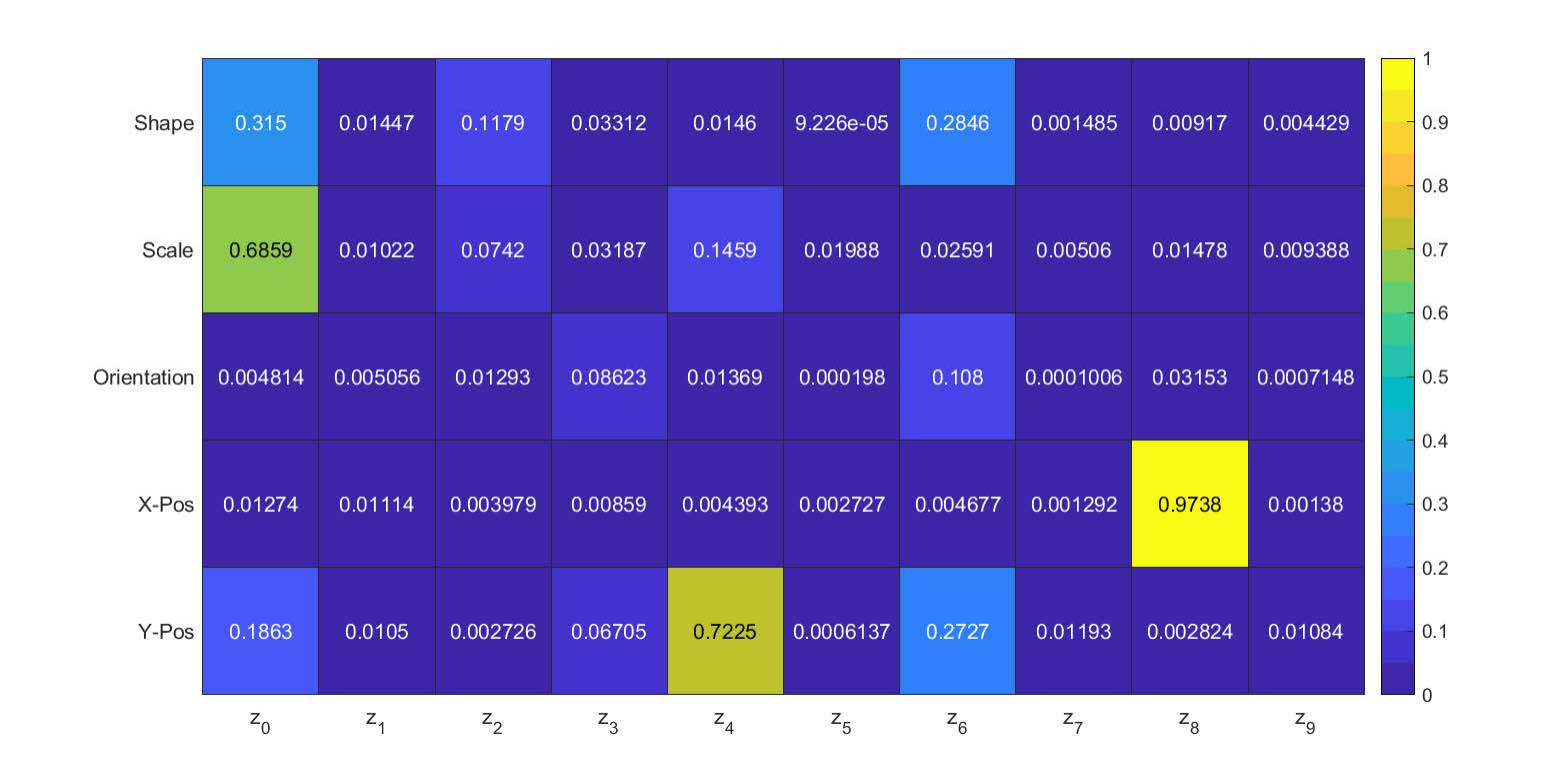}
}
\\
\centering
\subfloat[VAE+EC\label{fig:corr110}]
{
\includegraphics[width=0.5\linewidth, trim=40 0 40 0]{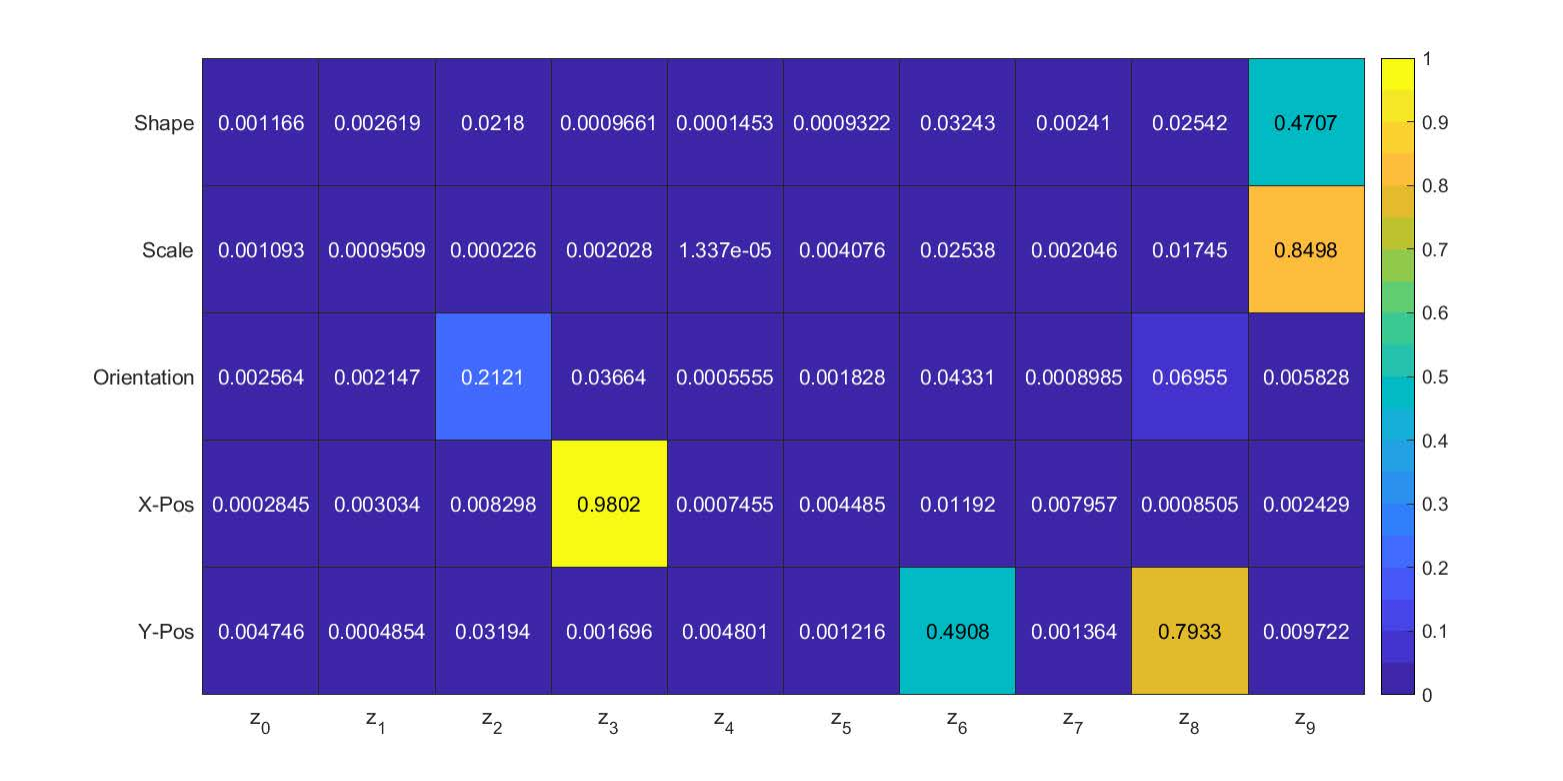}
}
\subfloat[VAE+EC+LB\label{fig:corr111}]
{
\includegraphics[width=0.5\linewidth, trim=40 0 40 0]{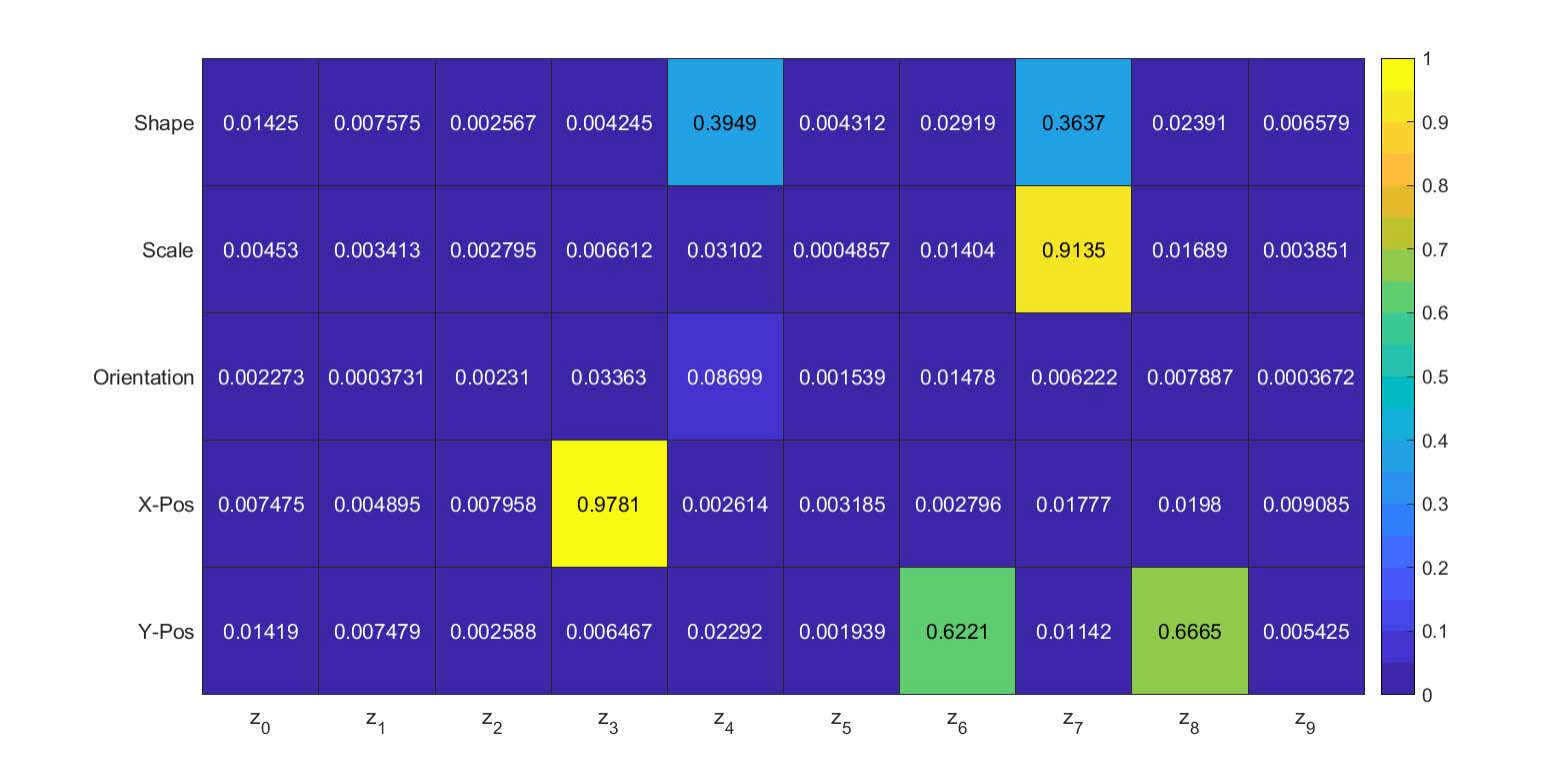}
}
\caption{Cross correlation between latent variables and labeled features in the vanilla VAE models. 
The vertical axis is the labels of the dSprites in Table~\ref{table:dSpritesLabels}. 
The horizontal axis is the latent variables from $\mathbf{z}_0$ to $\mathbf{z}_9$. 
The color of each box represents the absolute value of correlation between the label and latent variable. 
The higher value with lighter color indicates the latent variable encodes the more information of the certain label.}
\label{fig:corrb1}
\end{figure}

\begin{figure}[hbt] 
\centering
\subfloat[$\beta$-VAE\label{fig:corr400}]{
\includegraphics[width=0.5\columnwidth, trim=40 0 40 0]{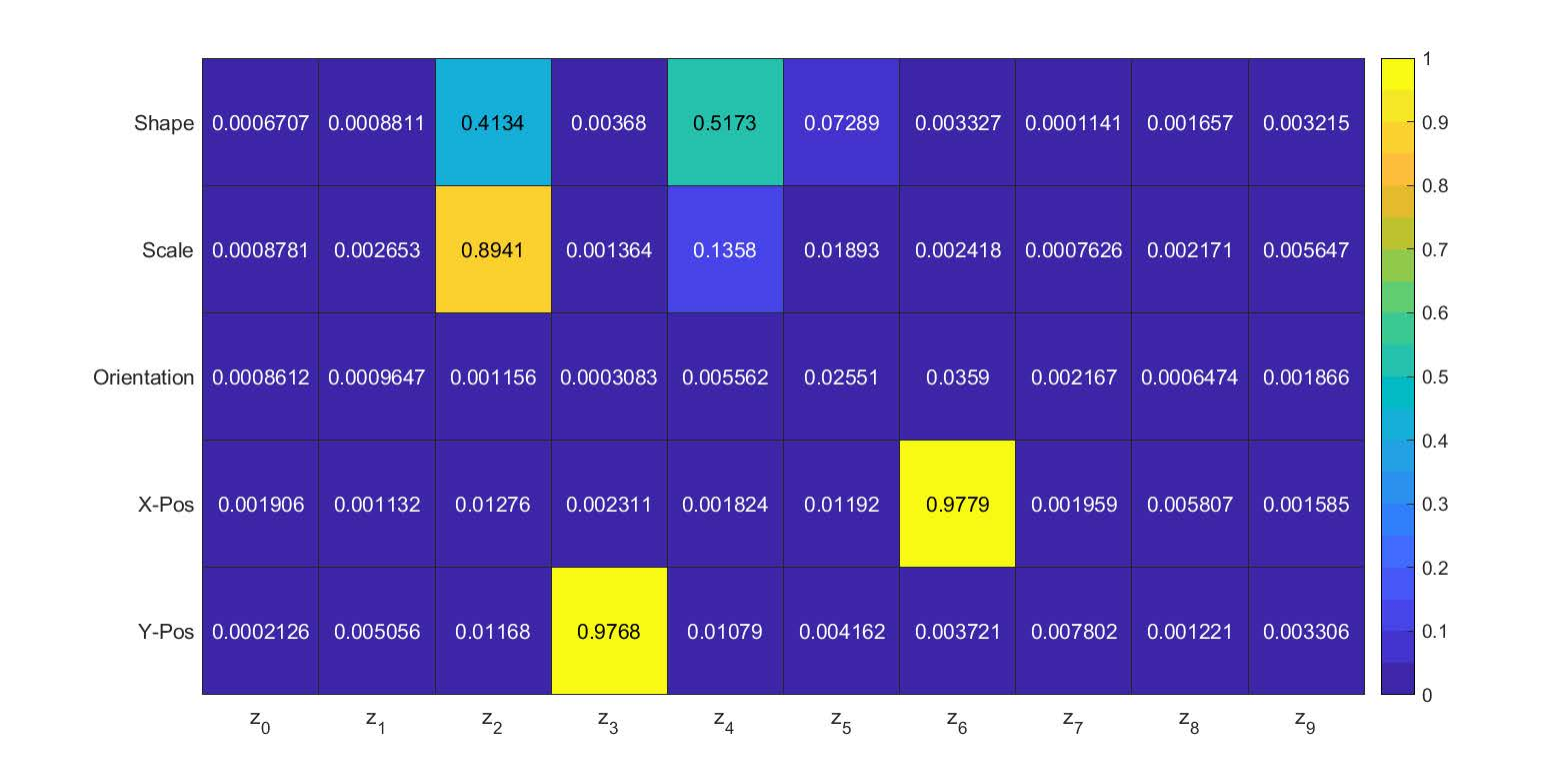}}\\
\centering
\subfloat[$\beta$-VAE+EC*\label{fig:corr40360}]{
\includegraphics[width=0.5\columnwidth, trim=40 0 40 0]{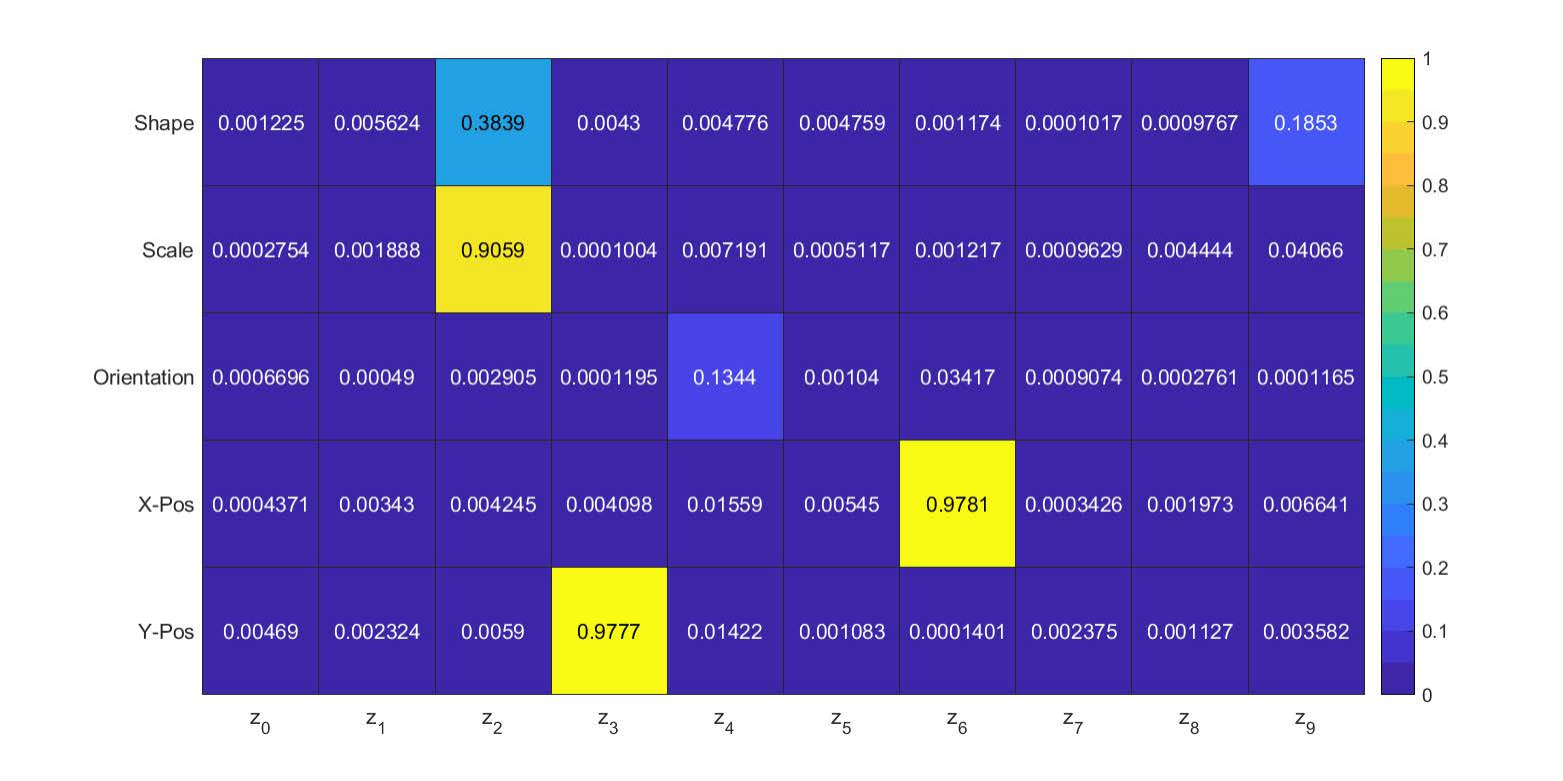}}
\subfloat[$\beta$-VAE+EC*+LB\label{fig:corr40361}]{
\includegraphics[width=0.5\columnwidth, trim=40 0 40 0]{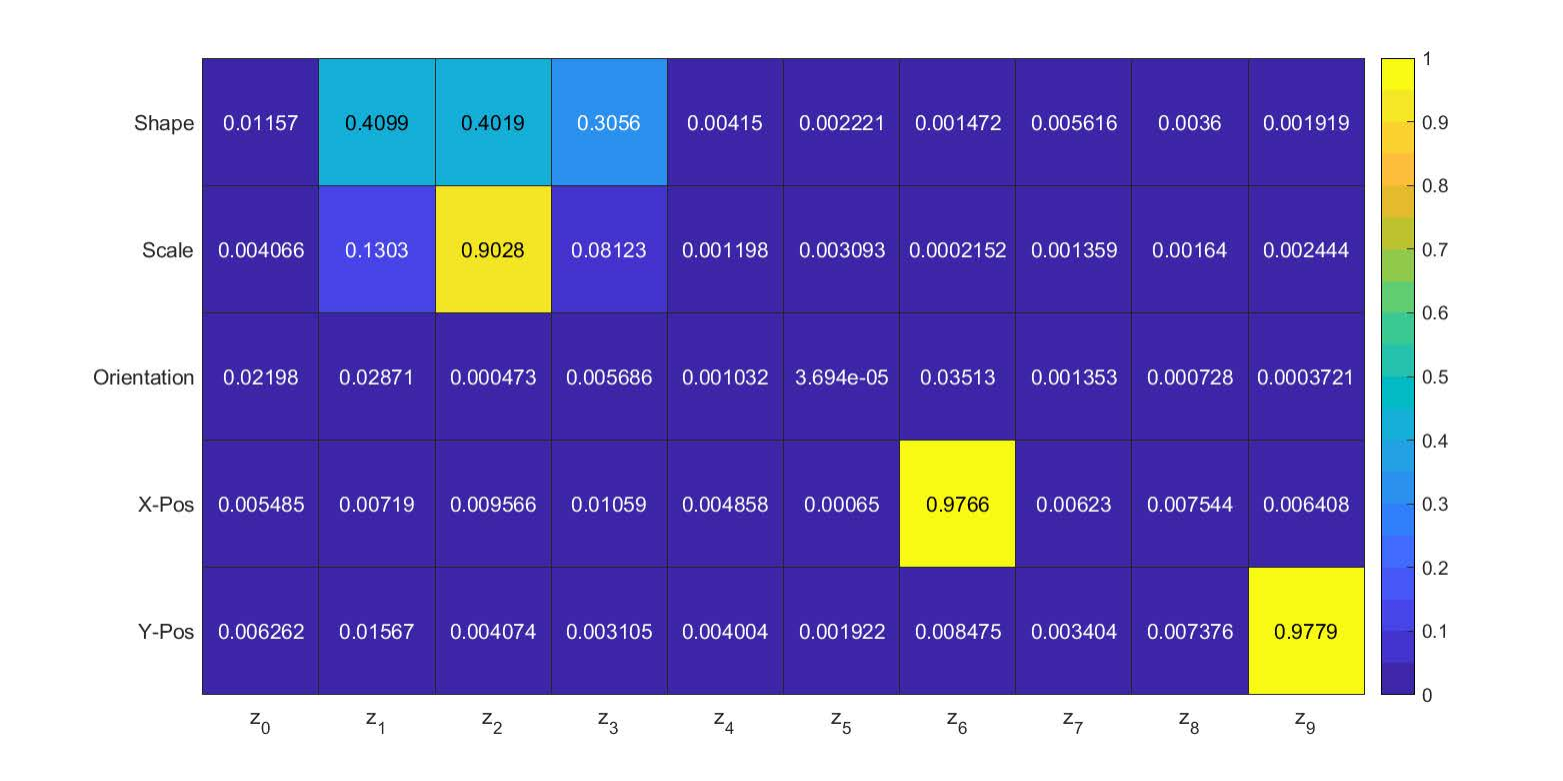}}\\
\centering
\subfloat[$\beta$-VAE+EC\label{fig:corr410}]{
\includegraphics[width=0.5\columnwidth, trim=40 0 40 0]{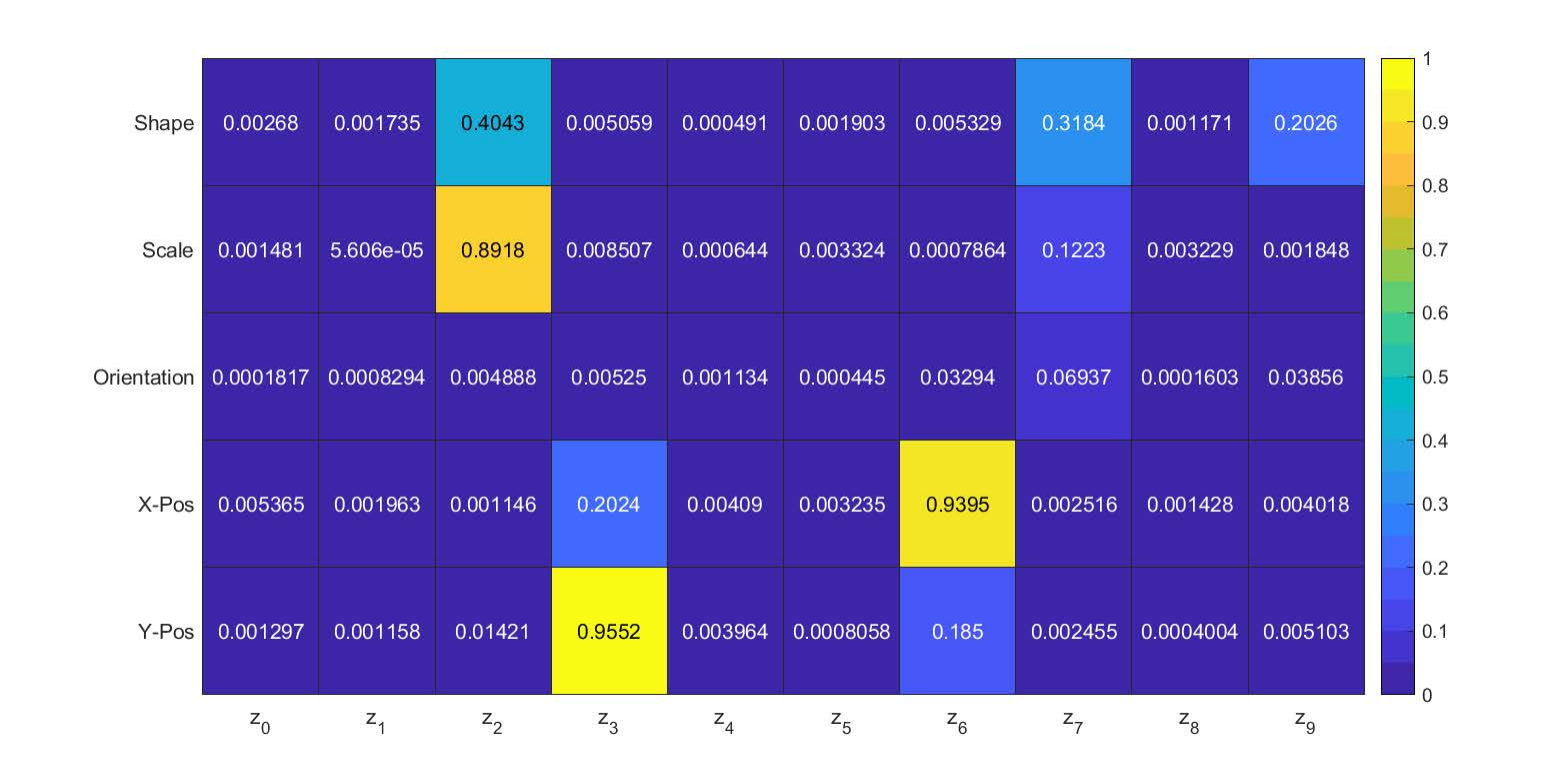}}
\subfloat[$\beta$-VAE+EC+LB\label{fig:corr411}]{
\includegraphics[width=0.5\columnwidth, trim=40 0 40 0]{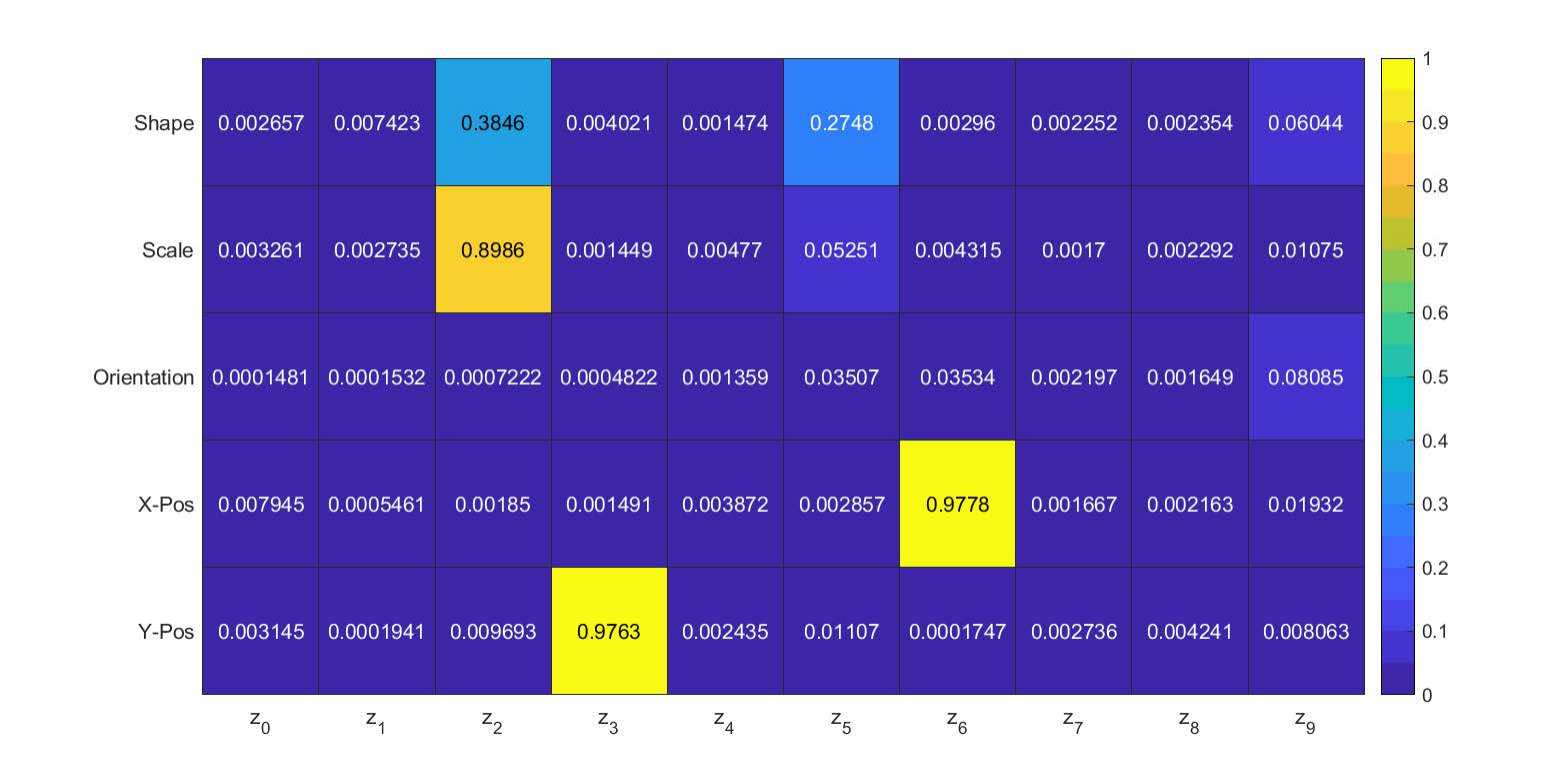}}
\caption{Cross correlation between latent variables and labeled features in the $\beta$-VAE Models. 
}
\label{fig:corrb4}
\end{figure}

The correlation scores are computed from the cross correlations in Fig.~\ref{fig:corrb1} and Fig.~\ref{fig:corrb4}.
We have included correlation plots in Fig.~\ref{fig:corrb1} to compare the relationship between the labels and latent variables in the vanilla VAEs.
The correlation plot in Fig.~\ref{fig:corr100} revealed 
that the vanilla VAE did not exhibit strong dependence 
between the labels and the latent variable in most cases.
However, our proposed models VAE+EC (Fig.~\ref{fig:corr110})
and VAE+EC+LB (Fig.~\ref{fig:corr111}) showed a more pronounced relationship 
between the labels and the corresponding latent variables, 
except for the shape and scale labels. 
Interestingly, the $Y$-position label in Figs.~\ref{fig:corr110} and~\ref{fig:corr111} corresponded to $\mathbf{z}_6$ and $\mathbf{z}_8$, respectively. 
It is worth noting that in most cases, 
each latent variable in our models corresponded only to one label.

Figure~\ref{fig:corrb4} presented the correlation plot 
in the $\beta$-VAE models, 
and we compared it with the plot in Fig.~\ref{fig:corrb1}. 
We observed that the correlation patterns in Fig.~\ref{fig:corrb4} were more consistent across all plots. 
All models in Fig.~\ref{fig:corrb4} showed lower correlations for the orientation label.
However, our models in Figs.~\ref{fig:corr40360}, ~\ref{fig:corr410}, and~\ref{fig:corr411} exhibited 
a more distinguishable correlation pattern for the orientation label, 
as they had an independent latent variable that corresponded to it.
This is why our proposed approach achieved higher FactorVAE scores for the orientation label as shown in Fig.~\ref{fig:betafactorDM}. 
Orientation is a special case.
Although the label takes 40 nominal values, the rotational symmetry of the shapes
makes many of those values indistinguishable in the rendered image, so the
effective entropy of the factor is far lower than the nominal count suggests and
its correlation with any latent variable is correspondingly weak.
Section~\ref{sec:irregular} examines this in detail.

\subsection{Demonstration of the Optimization of Latent-space Dimensionality}
\begin{figure}[htbp]
\centering
\includegraphics[width=\linewidth]{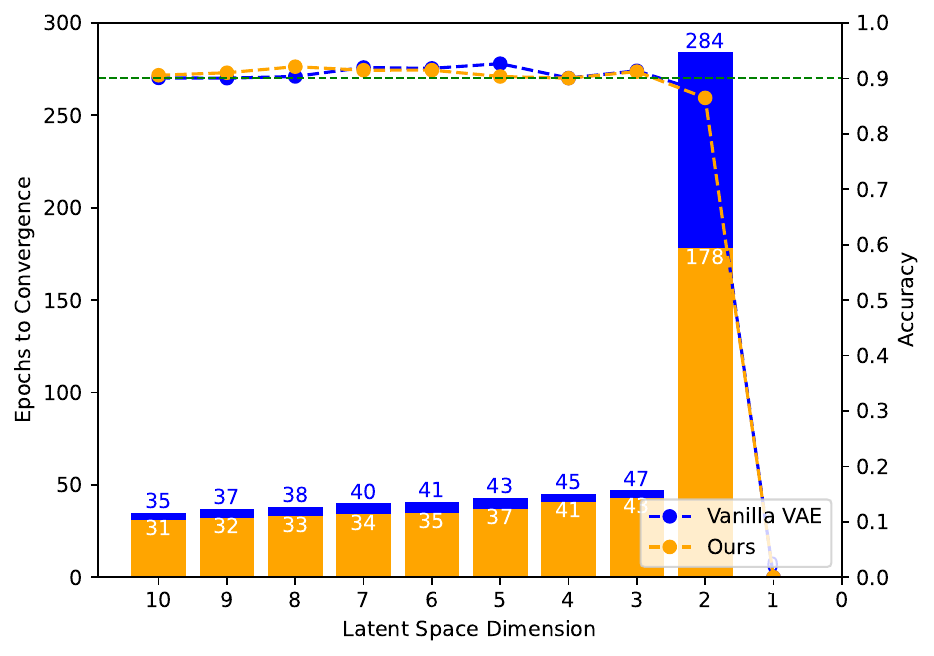}
\caption{MNIST latent-space dimension versus classifier accuracy and convergence epochs.
The dash curves indicate the changes of the accuracy along the decreasing of the dimension.
The green dash line is the threshold $\alpha = 0.9$.
The bar indicates the epoch number when the classifier converges.}
\label{fig:optdim}
\end{figure}

In this experiment, our objective is to minimize the dimensionality 
of the latent space and use the latent representation to train 
a MLP classifier on the MNIST dataset. 
Figure~\ref{fig:optdim} shows the number of epochs to convergence and the
classifier accuracy at each dimensionality.
Our model was a vanilla VAE with the EC ($\gamma=1$), compared against the same
procedure applied to a vanilla VAE without it.
The EC converged faster at every dimensionality: between 9\% and 15\% fewer
epochs from ten dimensions down to three, and 37\% fewer at two dimensions
($178$ against $284$ epochs), where the classifier is under the most pressure.
Both models then failed to maintain the accuracy threshold and halted the
reduction at two dimensions.
Within this setup---this classifier, this threshold $\alpha=0.9$, and this
dataset---two dimensions is therefore the smallest latent space the downstream
task tolerates.
We stress that this figure is a property of the task and the threshold rather
than of MNIST: a different classifier or a different $\alpha$ will return a
different value.

\begin{figure}[htbp]
\centering
\includegraphics[width=\linewidth]{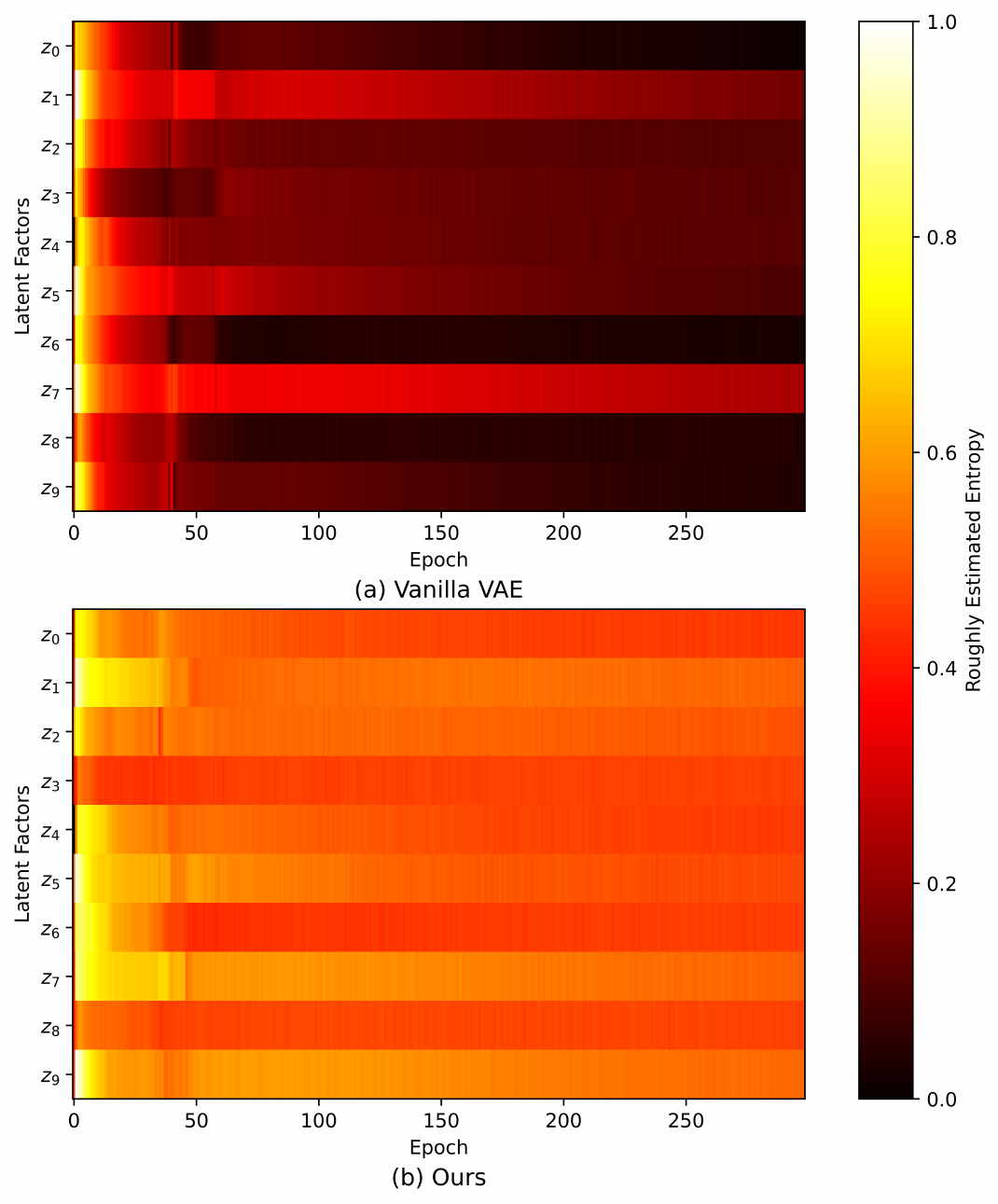}
\caption{Heatmap of normalized latent-variable entropy along training epochs.}
\label{fig:Zheatmap}
\end{figure}

In Fig.~\ref{fig:Zheatmap}, the heatmap displays the latent-variable entropies throughout the training epochs. 
The entropy, estimated using Algorithm~\ref{alg:EstEntropy}, indicates that our proposed encoder maintained a higher entropy throughout training due to the imposed EC. 
The entropy of our encoder exhibited a smoother transition from the beginning to the end of the training process, as shown in Fig.~\ref{fig:Zheatmap}b. 
This stability in entropy ensured faster convergence of the classifier and enhanced overall training stability.

\section{Analysis of the Distributions of Latent Variables}\label{sec:discussion}
The EC on latent variables operates as a penalty term within the loss function
that directs the optimizer to raise latent-variable entropy.
In this section we examine the effect of that term on the distributions of the
latent variables themselves, which is where the mechanism behind the numbers in
Section~\ref{sec:experiment} becomes visible.

\subsection{Entropy of Latent Variables}
\begin{figure}[htb]
\centering
\subfloat[Vanilla VAE\label{fig:hist100}]
{
\includehist{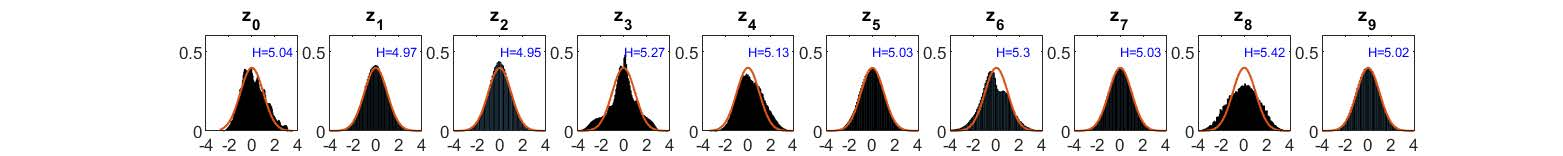}
}
\\
\centering
\subfloat[VAE+EC\label{fig:hist110}]
{
\includehist{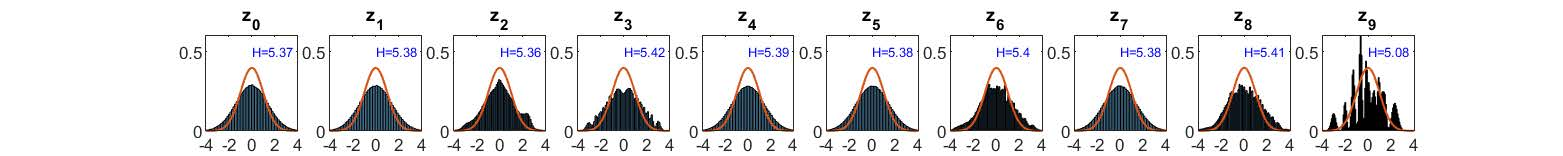}
}
\\
\centering
\subfloat[VAE+EC+LB\label{fig:hist111}]
{
\includehist{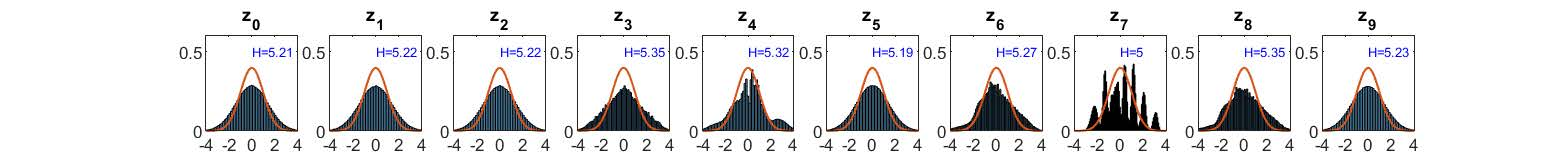}
}
\caption{Vanilla VAE latent-variable histograms of the dSprites dataset.
In each latent variable's histogram, 
the black shade is the histogram and the red curve is a normal distribution PDF $\mathcal{N}(0,1)$. 
The entropy of each latent variable is marked 
on the top right corner of each histogram. 
When the black shade and red curve are further, 
it indicates the latent variable has higher variance and higher entropy.}
\label{fig:Histb1}
\end{figure}

\begin{figure}[htb]
\centering
\subfloat[$\beta$-VAE\label{fig:hist400}]
{
\includehist{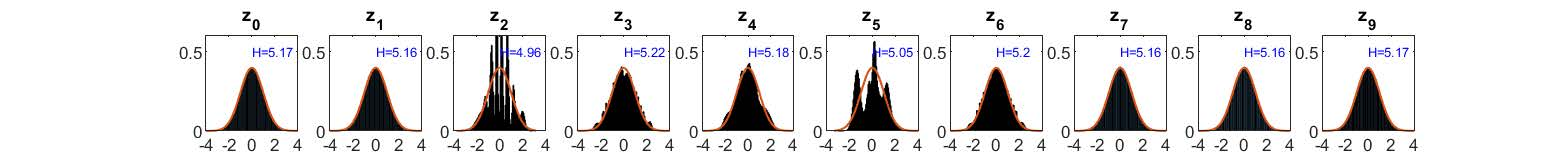}
}
\\
\centering
\subfloat[$\beta$-VAE+EC*\label{fig:hist40360}]
{
\includehist{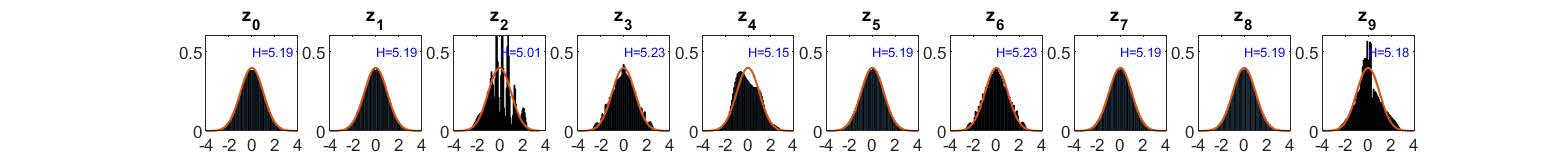}}
\\
\centering
\subfloat[$\beta$-VAE+EC*+LB\label{fig:hist40361}]
{
\includehist{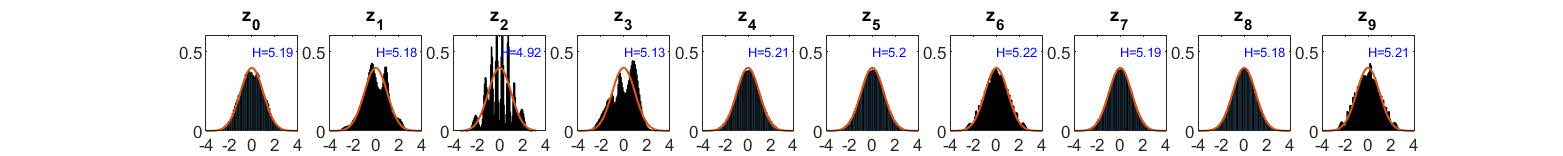}
}
\\
\centering
\subfloat[$\beta$-VAE+EC\label{fig:hist410}]
{
\includehist{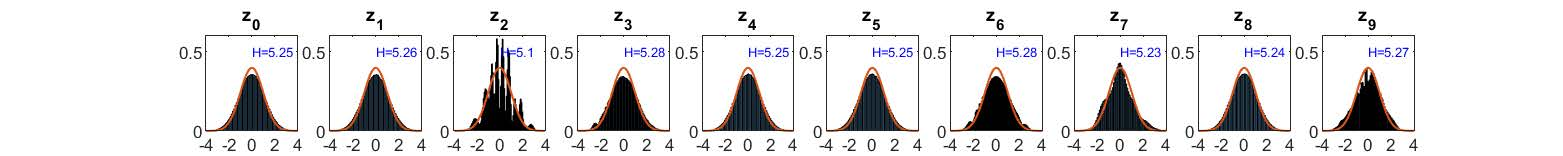}
}
\\
\centering
\subfloat[$\beta$-VAE+EC+LB\label{fig:hist411}]
{
\includehist{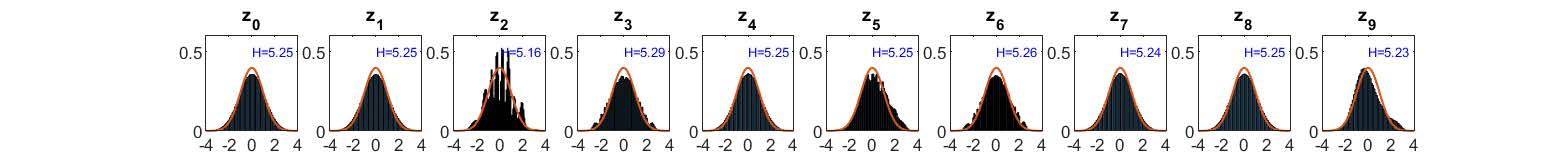}
}
\caption{$\beta$-VAE latent-variable histograms of the dSprites dataset. 
The figure setup is identical to Fig.~\ref{fig:Histb1}.
}
\label{fig:Histb4}
\end{figure}

Figures~\ref{fig:Histb1} and~\ref{fig:Histb4} show the histogram and estimated entropy of the latent variables in vanilla VAE and $\beta$-VAE models, respectively.
Compared to the vanilla VAE in Fig.~\ref{fig:hist100}, 
our models exhibited a wider distribution. 
Our approach favored encoding as much information as possible of 
a label in a latent variable. 
Recalling the correlation plots from Figs.~\ref{fig:corrb1} and~\ref{fig:corrb4}, 
all labels correlated with at least one latent variable in our models with a high value compared to the vanilla VAE and $\beta$-VAE.

When the entropy of a label feature was lower 
and the classes were distributed discretely, 
our methods favored combining these discrete features 
and encoding them into one latent variable. 
As seen in Figs.~\ref{fig:corr110}, ~\ref{fig:corr111}, ~\ref{fig:corr40360}, ~\ref{fig:corr410}, and~\ref{fig:corr411}, 
the shape and scale labels are merged in our models 
and encoded into a single latent variable, $\mathbf{z}_2$, 
for which the number of distinct label combinations is $3\times6=18$ rather than
the $3$ and $6$ of the individual labels. 
The histograms in Fig.~\ref{fig:Histb4} corroborate this: the distribution of
$\mathbf{z}_2$ shows many local peaks, one group per combination, arranged so that
the envelope still approximates the normal prior.
We describe this merging as a property that the EC preserves and makes legible in
the correlation structure, rather than as one it introduces; a comparable pattern
is discernible in the $\beta$-VAE baseline of Fig.~\ref{fig:corr400}, and what
distinguishes our models is the strength and the isolation of the association
rather than its existence.

On the other hand, when a label's entropy was high and  
the classes in the feature were continuous, 
our VAE models separated the data feature 
and embedded it into more than one latent variable.
As seen in Figs.~\ref{fig:corr110} and~\ref{fig:corr111}, 
the $Y$-position label was encoded in the latent variables $\mathbf{z}_6$ and $\mathbf{z}_8$.

\subsection{Latent Variables in Gaussian Mixture Distribution}
By introducing the $\beta$ parameter in the VAE loss function, 
$\beta$-VAE models encouraged the latent variables to have 
a more locally centralized distribution as seen in Figs.~\ref{fig:Histb1} and~\ref{fig:Histb4}. 
This property is advantageous when encoding continuous features 
such as the $X$ and $Y$ positions, resulting in disentangled normal distributions such as $\mathbf{z}_3$ and $\mathbf{z}_6$ in Fig.~\ref{fig:Histb4}. 
However, when encoding discrete features such as shape and scale, 
the normal distribution of the latent variable was no longer maintained. 
Instead, the latent variable formed a Gaussian mixture distribution 
with multiple local means such as $\mathbf{z}_2$ in Fig.~\ref{fig:Histb4}. 
It is worth noting that the histograms show the aggregate posterior
$q_\phi(\mathbf{z})=\mathbb{E}_{\mathbf{x}}[q_\phi(\mathbf{z}|\mathbf{x})]$, that
is, the marginal over the data, and not any individual per-input posterior.
A multi-modal aggregate posterior is therefore compatible with every
per-input posterior being Gaussian, which is exactly what a mixture over discrete
factor values produces.

\subsection{Encoding Irregular Data Features}\label{sec:irregular}
Our constraint was effective in encoding the orientation label of the dSprites
dataset, which is the clearest case of an irregularly distributed factor.
Both our results and previous work~\cite{burgess2018understanding} identify
orientation as the most difficult feature to encode.
The reason is a mismatch between the label and the image it produces.
Although dSprites samples orientations uniformly over $[0,2\pi]$, a shape with
rotational symmetry maps distinct orientation labels onto identical images: the
square is invariant under rotations by multiples of $\pi/2$ and the ellipse under
multiples of $\pi$, while the heart has no rotational symmetry at all.
The number of distinguishable orientations therefore differs across the three
shapes, the orientation factor is effectively non-uniform over the rendered
images even though the label is uniform, and its correlation with the other
factors is weak.

Our fine-tuned model significantly improved the $\beta$-VAE 
disentanglement score as shown in Fig.~\ref{fig:betabetaDM}. 
Comparing, in Fig.~\ref{fig:corrb4}, the latent variable most strongly correlated
with orientation in each model, we found that every EC variant reached a higher
correlation with the orientation label than the $\beta$-VAE baseline. 
The latent variable $\mathbf{z}_5$ in Fig.~\ref{fig:hist400}, 
which had the highest correlation with the orientation label 
in the $\beta$-VAE model, has a Gaussian mixture-like distribution.

In our implementation, all models were trained on a balanced dataset, 
which likely resulted in uniformly distributed data features.
However, the irregular distribution of the orientation label highlighted 
the challenge of working with unbalanced datasets, 
which could be a subject for future work.

\subsection{Induction Bias}
Recent research~\cite{Locatello2020} has demonstrated that induction bias in unsupervised disentanglement is significantly influenced by random seed selection. 
A judicious choice of random seed can lead to perfect disentanglement and distinctive patterns in latent-variable encoding. 
This phenomenon is evident in our implementation, as depicted in Figs.~\ref{fig:corrb1} and~\ref{fig:corrb4},
where different random seeds were used for the vanilla VAE and $\beta$-VAE models, 
resulting in markedly different encoding patterns.

We trained every model under ten random seeds.
For each family we then report the single seed that produced the lowest
reconstruction error: seed~1 for the vanilla-VAE family
(Tables~\ref{table:vanillaresults} and~\ref{tbl:vanillametrics}) and seed~10 for
the $\beta$-VAE family (Tables~\ref{table:beta4results}
and~\ref{tbl:betametrics}).
Within a family all models share a seed and are directly comparable; across
families they are not, which is why we report the two families separately
throughout Section~\ref{sec:experiment} rather than in a single table.
This is a best-case rather than an expected-case protocol, and we treat it as a
limitation below.

\subsection{Limitations}\label{sec:limitations}
Four limitations of the present study deserve statement.

First, the seed protocol described above reports the best of ten runs rather than
a mean over them, and the two model families were selected under different seeds.
The tables therefore support statements about which configuration can perform best
but not about expected performance or statistical significance.

Second, the comparison covers the vanilla VAE and $\beta$-VAE only.
Methods that target the same trade-off directly---FactorVAE~\cite{Kim_Mnih},
$\beta$-TCVAE~\cite{Chen_Li_Grosse_Duvenaud_2019}, and the importance-weighted and
Wasserstein variants surveyed in Section~\ref{sec:relatedwork}---are discussed but
not run, so the evidence that the EC is preferable to them remains indirect.
Relatedly, disentanglement is assessed with two classifier-based metrics that
disagree about our models, and two metrics cannot adjudicate between themselves.
A third measure of different construction, such as the informativeness and
disentanglement axes of the DCI
framework~\cite{Eastwood_Williams_2018}, would show whether the disagreement is a
property of the learned representation or of the metrics, and this is the clearest
gap in the present evidence.

Third, the entropy estimate of Algorithm~\ref{alg:EstEntropy} is a histogram
plug-in estimator.
It is biased downwards for a fixed sample size, its value depends on the bin
count, and it is applied per dimension, so it cannot detect the dependence
between latent variables that Eq.~\eqref{eqn:TC} relies on being small.
The constraint is imposed on a quantity whose scale differs from that of the
differential-entropy target $\delta$, as noted in Section~\ref{sec:Optlatent}.

Fourth, the weight filter was evaluated on one downstream task, with one
classifier architecture and one threshold, $\alpha=0.9$.
The dimensionality it returns is a property of that combination, and its greedy
schedule never reinstates an eliminated dimension, so the reported $J$ is feasible
rather than provably minimal.
The $1/J$ weighting also rescales the classifier input at every reduction step, so
the epoch counts of Fig.~\ref{fig:optdim} reflect the reduced dimensionality and
that rescaling together; the effect is common to both compared models, but
isolating it would require an ablation with fixed unit weights.

\section{Conclusion}\label{sec:conclusion}
This paper approached the optimal latent space of a VAE from two directions at
once---the capacity of the individual latent variables and the dimensionality of
the space they span---and treated both within a single soft-constrained
formulation.
The entropy-based constraint gives the capacity requirement an explicit
representation in the objective, and we related it to the mutual information
between the latent code and the generative factors: the entropy of the code
bounds that mutual information, with the residual term determined by the encoder
variances and hence controlled by the existing KL regularizer.
On dSprites the constraint raised the aggregate activation score by 43--62\% over
a vanilla VAE and by up to 17\% over a $\beta$-VAE, improved our
correlation-based score by 9--14\%, and produced the highest FactorVAE score
among the $\beta$-VAE variants.
It did not improve the $\beta$-VAE metric, and its effect on reconstruction error
depends on $\gamma$; the best configuration reduced that error by 38\% while the
worst increased it.

The weight-filter method exploits the slack of the soft constraint to remove the
latent dimensions that satisfy it least well.
On MNIST it reduced the dimensionality consumed by a downstream classifier from
ten to two while holding accuracy above the threshold, converging in 37\% fewer
epochs at that dimensionality than the same procedure without the constraint, and
identified two dimensions as the point at which the threshold can no longer be
met for this task.

The analysis of the latent distributions showed why the two effects arise
together: low-entropy discrete factors are merged into a single latent variable
whose aggregate posterior becomes a Gaussian mixture, whereas high-entropy
continuous factors are distributed across several variables that remain close to
Gaussian.

Several directions follow from the limitations of
Section~\ref{sec:limitations}.
A seed-averaged comparison against FactorVAE and $\beta$-TCVAE would establish
where the EC sits among methods that target the same trade-off.
Replacing the histogram estimator with a lower-variance alternative, or
constraining the joint entropy directly rather than the marginals, would tighten
the link between the constraint and the quantity it is meant to control.
Finally, the behavior observed on the orientation factor suggests that
unbalanced or symmetry-degenerate factors are where an entropy-based constraint
has the most to offer, and that is where we intend to take it next.

\bibliographystyle{IEEEtran}

\bibliography{mybibfile}

\end{document}